\documentclass{article}
\usepackage{latexsym}
\usepackage{amsmath, amssymb, amsfonts, amsthm}
\usepackage{epsfig}
\usepackage{stmaryrd}
\usepackage{multicol}
\usepackage{pdfsync}
\usepackage{dsfont}
\usepackage{comment}
\usepackage{algorithmic}
\usepackage{algorithm}
\usepackage{cite}
\usepackage[font=footnotesize,format=plain,labelfont=bf]{caption}
\usepackage{psfrag}
\usepackage{subfig}
\usepackage[dvipsnames]{xcolor}
\usepackage{graphics}
\usepackage{enumitem}
\usepackage{mathtools}
\usepackage{fullpage}
\usepackage{booktabs}
\usepackage{multirow}
\usepackage{siunitx}

\graphicspath{{./Figures/}}

\newcommand{\DNSsub}{{\text{\textsc{dns}}}}
\newcommand{\LESsub}{{\text{\textsc{les}}}}

\newcommand{\norm}[1]{\left\lVert #1 \right\rVert}

\newcommand{\fbw}{\ol{\mathbf{w}}}
\newcommand{\fbU}{\ol{\mathbf{U}}}
\newcommand{\hbu}{\widetilde{\mathbf{u}}}
\newcommand{\fbu}{\ol{\mathbf{u}}}
\newcommand{\fDelta}{\ol{\Delta}}

\newcommand{\bbf}{\mathbf{f}}
\newcommand{\bh}{\mathbf{h}}
\newcommand{\bu}{\mathbf{u}}
\newcommand{\bx}{\mathbf{x}}
\newcommand{\bw}{\mathbf{w}}

\newcommand{\bV}{\mathbf{V}}
\newcommand{\btau}{\boldsymbol{\tau}}
\newcommand{\cD}{\mathcal{D}}
\newcommand{\cA}{\mathcal{A}}
\newcommand{\chA}{\widehat{\mathcal{A}}}
\newcommand{\eps}{\varepsilon}
\usepackage[colorinlistoftodos, color=blue!20!white, bordercolor=gray,
  textsize=tiny,textwidth=0.8in,disable]{todonotes}

\theoremstyle{definition}

\theoremstyle{remark}

\numberwithin{equation}{section}

\def \d {\mathrm{d}}

\newcommand{\pp}[2]{\frac{\partial #1}{\partial #2}}

\newcommand{\ol}[1]{\overline{#1}}

\newcommand{\wh}[1]{\widehat{#1}}

\definecolor{lc1}{HTML}{CE282A}
\definecolor{lc2}{HTML}{255BCB}
\definecolor{lc3}{HTML}{2DCE65}
\definecolor{lc4}{HTML}{A21D8E}
\definecolor{lc5}{HTML}{F1AB65}
\definecolor{lc6}{HTML}{89B6D4}

\DeclareMathOperator*{\argmin}{arg\,min}

\title{\Large DPM:  A deep learning PDE augmentation method \\ (with application to large-eddy simulation)}

\author{Jonathan B.~Freund\footnote{Mechanical Science \& Engineering and Aerospace Engineering, University of Illinois at Urbana--Champaign, jbfreund@illinois.edu}, Jonathan F.~MacArt\footnote{The Center for Exascale Simulation of Plasma-coupled Combustion, Coordinated Science Laboratory, University of Illinois at Urbana--Champaign,  jmacart@illinois.edu}, and Justin Sirignano\footnote{Department of Industrial \& Systems Engineering, University of Illinois at Urbana--Champaign, 
jasirign@illinois.edu} \footnote{The author list is alphabetical.}}

\date{\today}

\begin{document}

\maketitle

\begin{abstract}

Machine learning for scientific applications faces the challenge of limited data. We propose a framework that leverages \textit{a priori} known physics to reduce overfitting when training on relatively small datasets. A deep neural network is embedded in a partial differential equation (PDE) that expresses the known physics and learns to describe the corresponding unknown or unrepresented physics from the data. Crafted as such, the neural network can also provide corrections for erroneously represented physics, such as discretization errors associated with the PDE's numerical solution. Once trained, the deep learning PDE model (DPM) can make out-of-sample predictions for new physical parameters, geometries, and boundary conditions.

%This formulation can be viewed as optimizing the PDE itself, so estimating the embedded neural network requires optimizing over the entire PDE, which itself is a function of the neural network.

Our approach optimizes over the functional form of the PDE. Estimating the embedded neural network requires optimizing over the entire PDE, which itself is a function of the neural network. Adjoint partial differential equations are used to efficiently calculate the high-dimensional gradient of the objective function with respect to the neural network parameters. A stochastic adjoint method (SAM), similar in spirit to stochastic gradient descent, further accelerates training.

The approach is demonstrated and evaluated for turbulence predictions using large-eddy simulation (LES), a filtered version of the Navier--Stokes equation containing unclosed sub-filter-scale terms.  High-fidelity direct numerical simulations (DNS) of decaying isotropic turbulence provide the training and testing data. The DPM outperforms the widely-used constant-coefficient and dynamic Smagorinsky models, even for filter sizes so large that these established models become qualitatively incorrect. It also significantly outperforms \textit{a priori} trained models, which do not account for the full PDE.  For comparable accuracy, the overall cost is reduced. Simulations of the DPM are accelerated by efficient GPU implementations of network evaluations. Measures of discretization errors, which are well-known to be consequential in LES, suggest that the ability of the training formulation to correct for these errors is crucial to its success. A relaxation of the discrete enforcement of the divergence-free constraint is also considered, instead allowing the DPM to approximately enforce incompressibility physics.
 
\end{abstract}

\section{Introduction}
\label{Intro}
It is well-understood that machine learning in science and engineering is challenged by limited availability of data. Experiments can be expensive and time-consuming, and some quantities are difficult or impossible to measure with current techniques. Similarly, high-fidelity numerical simulations are computationally costly or even infeasible for many applications. To reduce overfitting on relatively small datasets, our framework leverages the \textit{a priori} known and representable physics by embedding a deep neural network $\bh_{\theta}(u, v, w)$ in a PDE description of the known and represented physics.  This yields the class of PDE models:
\begin{eqnarray}
\frac{\partial \mathbf{u} }{\partial t} (t, \mathbf{x})  = \underbrace{ \bbf_{\nu} \big{(} \mathbf{u}(t,\mathbf{x}), \mathbf{u}_\mathbf{x}(t,\mathbf{x}), \mathbf{u}_{\mathbf{x} \mathbf{x}}(t, \mathbf{x}) \big{)}}_{\textrm{Known and represented physics}} + \underbrace{\bh_{\theta} \big{(} \mathbf{u}(t, \mathbf{x}), \mathbf{u}_\mathbf{x}(t, \mathbf{x}), \mathbf{u}_{\mathbf{x} \mathbf{x} }(t, \mathbf{x} ) \big{)} }_{\textrm{Unknown/unrepresented physics}}, 
\label{DPM}
\end{eqnarray}
where $\mathbf{u}(t, \mathbf{x}) = \big{(} u_1(t, \mathbf{x}), \ldots, u_d(t, \mathbf{x}) \big{)} \in \mathbb{R}^{d}$, $\mathbf{x} = (x_1, \ldots, x_q) \in \Omega \subset \mathbb{R}^q$, $\mathbf{u}_{\mathbf{x}}(t, \mathbf{x}) = \big{(} u_{1,\mathbf{x}} (t, \mathbf{x}), \ldots, u_{d,\mathbf{x}}(t, \mathbf{x}) \big{)} \in \mathbb{R}^{d \times q}$, and $\mathbf{u}_{\mathbf{x} \mathbf{x}}(t, \mathbf{x}) = \big{(} u_{1, x_1 x_1}(t, \mathbf{x}), \ldots, u_{1, x_q x_q}(t, \mathbf{x}), \ldots, u_{d,x_1 x_1}(t, \mathbf{x}), \ldots, u_{d,x_q x_q}(t, \mathbf{x})  \big{)} \in \mathbb{R}^{d \times q}$. The term $\bbf_{\nu}: \mathbb{R}^{(2q +1) d} \rightarrow \mathbb{R}^d$ represents the known physics in the PDE, with $\nu$ the physical and scenario parameters that specify the application.  Generalization to mixed-second-order and higher-order derivatives is straightforward. Appropriate initial and boundary conditions complete the PDE. The neural network $\bh_{\theta}: \mathbb{R}^{(2q +1) d} \rightarrow \mathbb{R}^d$ will learn to describe the unknown or unrepresented physics (more precisely defined below) using the neural network parameters $\theta \in \mathbb{R}^{d_{\theta}}$, which are estimated from the data.

Two classes of unknown or unrepresented physics are anticipated. In the first, the governing equations are too computationally expensive to solve. In this case, deep learning can be used to develop a reduced model. For the Navier--Stokes turbulence example that we consider, the reduced model includes a sub-grid-scale stress closure that attempts to represent the physical effects of small unresolved turbulence scales on those represented in the large-eddy simulation (LES).  The second case occurs when important physics is omitted, either because it is deemed too costly (for example, multi-component diffusion in reacting flows) or if it is truly unknown. In general, unknown or unrepresented physics will diminish the PDE model's accuracy. In such cases, the selected PDE model is incomplete although not incorrect, and thus it can be used as a building block for the development of machine learning models. The proposed deep learning formulation is designed to augment these governing equations based on additional though limited high-fidelity simulation or experimental data.

Equation (\ref{DPM}) is designated a deep learning PDE model (DPM). The model parameters $\theta$ are estimated from trusted data $\mathbf{V}(t, \mathbf{x}; \nu)$, which are assumed to be available at certain times $t_1, \ldots, t_{N_t}$ and for certain physical or scenario parameters $\nu_1, \ldots, \nu_M$. These are used to numerically estimate a neural network $\bh_{\theta}$ that generates the solution $\mathbf{u}$ to (\ref{DPM}) that most closely matches the trusted data $\mathbf{V}$. This is done by minimizing the objective function
\begin{equation}
L(\theta) = \sum_{m = 1}^M \sum_{n=1}^{N_t} \int_{\Omega} \norm{ \mathbf{u}(t_n, \mathbf{x}; \nu_m) - \mathbf{V}(t_n, \mathbf{x}; \nu_m) } d \mathbf{x},
\label{ObjectiveFunction}
\end{equation}
where we explicitly denote the dependence of $\mathbf{u}$ on $\nu$ via the notation $\mathbf{u}(t, \mathbf{x}; \nu)$.

Minimizing (\ref{ObjectiveFunction}) is challenging since it is a function of the PDE (\ref{DPM}), which in turn is a nonlinear function of the neural network parameters $\theta$. A stochastic adjoint method (SAM) is proposed for computational efficiency (Section~\ref{SAM}). It accelerates the optimization by calculating gradients with respect to the neural network parameters using adjoint PDEs. Although the dimension of $\theta$ can be large (easily $\gtrsim 10^5$ parameters), the number of adjoint PDEs matches the number of PDEs $d$ in (\ref{DPM}). 

Once trained on available datasets, the deep learning PDE model can be used for out-of-sample predictions in new scenarios with different $\nu$ or different boundary conditions. As a demonstration, we evaluate the accuracy of the DPM for out-of-sample initial conditions and physical parameters for the challenging case of LES of turbulence. In LES, the nonlinear Navier--Stokes equations are filtered and, to accelerate numerical solutions, are solved on coarse grids that purposefully do not resolve the full range of turbulence scales (see Section~\ref{DNS}). This reduces the computational cost compared to full-resolution direct numerical simulation (DNS), making LES tractable for many problems of engineering interest. However, the filtering step introduces unclosed terms into the LES equations (Section~\ref{LES}) that correspond to unrepresented physics in~\eqref{DPM}. The accuracy of LES strongly depends upon accurately modeling this unrepresented physics.   

LES truncates the turbulence energy spectrum at or near the mesh resolution, and it is therefore inherently linked to discretization errors, which are largest for the smallest scales closest to the nominal truncation point. Some estimates of discretization errors are made in Section~\ref{DiscreteUnknownPhysics}. Coping with modeling errors concurrently with discretization error is a fundamental challenge in LES~\cite{Ghosal1996,Kravchenko1997,Chow2003}. The DPM training (Section~\ref{Results}) is able to correct for this discretization error, just as if it were incorrect or missing physics.  Overall, it recovers resolved spectra on significantly coarser LES meshes than established models (Section~\ref{sec.model_comparison}).  

The application of deep learning is an exciting direction in scientific computing~\cite{SirignanoSpiliopoulos2017_DGM,Menon1,Menon2,FreundDL2,PeymanGivi1,Karniadakis1,Karniadakis2,Ling1,Ling2,BergNystrom,PhysicsofFluidsLES,Brunton:2020,FreundMLPersp:2019}. Especially relevant recent efforts include closure models for the Reynolds-averaged Navier--Stokes (RANS) equations using an \emph{a priori} estimation method in which the optimization is de-coupled from the PDE~\cite{Ling1,Ling2}. Although this approach works for RANS, which is simpler in the sense that only a time-averaged quantity is sought for the closure of time-averaged PDEs, it performs poorly for LES (Section~\ref{Apriori}), probably due to the more complex link between the resolved turbulence dynamics and the necessarily unsteady sub-grid-scale energy transfer.  Wang~\textit{et al.}~\cite{PhysicsofFluidsLES} developed closure models for LES using \emph{a priori} training. 

Nonlinear physics with a strong coupling between the unknown terms (to be estimated) and the output variables motivates the proposed optimization over the entire PDE~\eqref{DPM}. Berg and Nystrom~\cite{BergNystrom} also developed an adjoint-based method for estimating a neural network coefficient function in a PDE from data, demonstrating it for a Poisson equation.  In that case, the function does not depend upon the PDE solution. In our case, adjoint equations are used to estimate the functional form of the nonlinear PDEs. In addition, Holland~\textit{et al.}~\cite{Duraisamy} recently used discrete adjoint methods to train a neural network for closure of RANS. 

The stochastic adjoint method to accelerate training is introduced in Section~\ref{SAM}. The DPM is then formulated for large-eddy simulation in Section~\ref{SGS}. Numerical discretization errors relevant to any LES are quantified and discussed in Section~\ref{DiscreteUnknownPhysics}. A specific numerical application to large-eddy simulation of isotropic turbulence is introduced and analyzed in Section~\ref{Results}, which includes discussion of the relatively poor performance of the corresponding \textit{a priori} approach (Section~\ref{Apriori}).

\section{Stochastic Adjoint Method} \label{SAM}

\subsection{Adjoint-based PDE gradient}

The objective function (\ref{ObjectiveFunction}) can be iteratively minimized using gradient descent,
\begin{eqnarray}
\theta_{k+1} = \theta_k - \alpha_k \nabla_{\theta}  L(\theta_k),
\label{GradientDescent}
\end{eqnarray}
where $\alpha_k$ is the so-called learning rate. This requires calculating the gradient $\nabla_{\theta} L(\theta)$, which is computationally challenging since $L(\theta)$ depends upon the solution of the PDE (\ref{DPM}). 
%na\"ive
A na\"ive approach would directly apply the gradient $\nabla_{\theta}$ to (\ref{DPM}) and derive a PDE for $\nabla_{\theta} \mathbf{u}$. However, the dimension of this PDE would match the dimension $d_{\theta}$ of the neural network parameters $\theta$, which is large (often over $10^5$ parameters), so such an approach is computationally intractable. Calculating $\nabla_{\theta} L$ using numerical differentiation with finite difference approximations would similarly require an intractable number of evaluations.  

The adjoint PDE provides a computationally-efficient method for calculating $\nabla_{\theta} L(\theta)$. It requires solving only $p$ PDEs per parameter update, which matches the dimension of $\bu$ in (\ref{DPM}).  For any deep learning model, $p$ will be many times smaller than $d_{\theta}$. The adjoint PDE can be viewed as a continuous-time PDE version of the usual neural network backpropagation algorithm.  For notational convenience we start by considering $M = 1$ and $d = q = 1$. The generalization to $M, d, q \geq 1$ is straightforward (see Section~\ref{MultiDimensionalAdjoint}). For $x \in \Omega$ and $0 \leq t < t_N$, 
\begin{equation}
\begin{split}
-\frac{\partial \wh{u}}{\partial t}  &=   \wh{u} \nabla_{u} f_{\nu_m} \big{(} u, u_x, u_{xx} \big{)}  \\
&-  \frac{\partial}{\partial x} \bigg{[} \wh{u} \nabla_{v} f_{\nu_m} \big{(} u, u_x, u_{xx} \big{)}  \bigg{]} \\
&+  \frac{\partial^2}{\partial x^2} \bigg{[} \wh{u} \nabla_{w} f_{\nu_m} \big{(} u, u_x, u_{xx} \big{)}  \bigg{]} \\
&+ \wh{u} \nabla_{u} h_{\theta} \big{(} u, u_x, u_{xx} \big{)} \notag \\
&- \frac{\partial}{\partial x} \bigg{[} \wh{u} \nabla_{v} h_{\theta} \big{(} u, u_x, u_{xx} \big{)} \bigg{]} \\
&+ \frac{\partial^2}{\partial x^2} \bigg{[} \wh{u} \nabla_{w} h_{\theta} \big{(} u, u_x, u_{xx} \big{)} \bigg{]} \\
&+ \sum_{n=1}^{N-1} \delta( t- t_n) \nabla_{u} \norm{ u(t_n, x; \nu_m) - V(t_n, x; \nu_m) }, 
\label{DPMdetAdjoint}
\end{split}
\end{equation}
with the final condition
\begin{eqnarray}
\wh{u}(t_N,x) &=& \nabla_{u} \norm{ u(t_N, x; \nu_m) - V(t_N, x; \nu_m) }. 
\end{eqnarray}
The gradient of the objective function (\ref{ObjectiveFunction}) is then
\begin{eqnarray}
\nabla_{\theta}  L(\theta) &=& \int_0^{t_N} \int_{\Omega}  \wh{u}(t,x) \nabla_{\theta}  h_{\theta} \big{(} u(t,x), u_x(t,x), u_{xx}(t, x) \big{)} \, dx dt.
\end{eqnarray}
For $M > 1$, $M$ such adjoint PDEs must be solved. The adjoint equations are generalized to $d, q \geq 1$ in Section~\ref{MultiDimensionalAdjoint}. 

\subsection{Stochastic optimization}

When $M$ and $N_t$ are large, the adjoint PDE method still requires significant computation time per gradient descent iteration (\ref{GradientDescent}). To accelerate training, we introduce a stochastic adjoint method (SAM), which randomly samples time intervals, allowing for a larger number of training iterations per computational time.  Its steps are:
\begin{itemize}
\item Select uniformly at random a scenario $m \in \{1, \ldots, M \}$ and time $n \in \{0, \ldots, N-1 \}$,
\item Solve (\ref{DPM}) on $[t_n, t_{n+1}]$ with initial condition $V_{(t_n, x; \nu_m)}$ taken from available data,
\item Evaluate the objective function
\begin{eqnarray}
J(\theta) =\int_{\Omega} \norm{ u(t_{n+1}, x; \nu_m) - V(t_{n+1}, x; \nu_m) }\, dx,
\label{ObjectiveFunction2}
\end{eqnarray}
\item Calculate $\nabla_{\theta} J(\theta)$ via its (time-reversed) adjoint PDE, which satisfies, for $x \in \Omega$ and $t_n \leq t < t_{n+1}$,
\begin{eqnarray}
-\frac{\partial \wh{u}}{\partial t}(t, x)  &=&   \wh{u}(t,x) \nabla_{u} f_{\nu_m} \big{(} u(t,x), u_x(t,x), u_{xx}(t,x) \big{)} - \frac{\partial}{\partial x} \bigg{[} \wh{u}(t,x) \nabla_{v} f_{\nu_m} \big{(} u(t,x), u_x(t,x), u_{xx}(t,x) \big{)}  \bigg{]} \notag \\
&+&  \frac{\partial^2}{\partial x^2} \bigg{[} \wh{u}(t,x) \nabla_{w} f_{\nu_m} \big{(} u(t,x), u_x(t,x), u_{xx}(t,x) \big{)}  \bigg{]} + \wh{u}(t,x) \nabla_{u} h_{\theta} \big{(} u(t,x), u_x(t,x), u_{xx}(t, x) \big{)} \notag \\
&-& \frac{\partial}{\partial x} \bigg{[} \wh{u}(t,x) \nabla_{v} h_{\theta} \big{(} u(t,x), u_x(t,x), u_{xx}(t, x) \big{)} \bigg{]} + \frac{\partial^2}{\partial x^2} \bigg{[} \wh{u}(t,x) \nabla_{w} h_{\theta} \big{(} u(t,x), u_x(t,x), u_{xx}(t, x) \big{)} \bigg{]} \notag \\
\wh{u}(t_{n+1},x) &=& \nabla_{u} \norm{ u(t_{n+1}, x; \nu_m) - V(t_{n+1}, x; \nu_m) }, \notag \\
\nabla_{\theta}  J(\theta) &=& \int_{t_n}^{t_{n+1}} \int_{\Omega}  \wh{u}(t,x) \nabla_{\theta}  h_{\theta} \big{(} u(t,x), u_x(t,x), u_{xx}(t, x) \big{)}\, dx dt,
\end{eqnarray}
\item Update the neural network parameters
\begin{eqnarray}
\theta_{k+1} = \theta_k - \alpha_k \nabla_{\theta}  J(\theta_k),
\end{eqnarray}
and
\item Repeat until a convergence criterion satisfied. 
\end{itemize}

\subsection{Adjoint equations in the multi-dimensional case} \label{MultiDimensionalAdjoint}
The adjoint equations for $d, q \geq 1$ follow from (\ref{DPMdetAdjoint}). Let $v_j$ and $w_j$ respectively be the arguments $\mathbf{u}_{x_j}$ and $\mathbf{u}_{x_j x_j}$ for the $\bbf$ and $\bh_{\theta}$ functions. Then, 
\begin{equation}
\begin{split}
-\frac{\partial \wh{\bu}}{\partial t}  &=  \nabla_{u} \bbf_{\nu_m} \big{(} \bu, \bu_{\bx}, \bu_{\bx \bx} \big{)}^{\top}  \wh{\bu}   \\
&- \sum_{j = 1}^q  \frac{\partial}{\partial x_j} \bigg{[}  \nabla_{v_j} \bbf_{\nu_m} \big{(} \mathbf{u}, \mathbf{u}_{\bx}, \mathbf{u}_{\mathbf{x} \mathbf{x} } \big{)}^{\top} \mathbf{\wh{u}}  \bigg{]}  \\
&+  \sum_{j=1}^q \frac{\partial^2}{\partial x_j^2} \bigg{[}  \nabla_{w_j} \bbf_{\nu_m} \big{(} \mathbf{u}, \mathbf{u}_{\mathbf{x}}, \mathbf{u}_{\bx \bx} \big{)}^{\top}  \mathbf{\wh{u}} \bigg{]}  \\
&+  \nabla_{u} \bh_{\theta} \big{(} \mathbf{u}, \mathbf{u}_{\mathbf{x}}, \mathbf{u}_{\bx \bx} \big{)}^{\top} \mathbf{ \wh{u}}  \\
&- \sum_{j=1}^q \frac{\partial}{\partial x_j} \bigg{[} \nabla_{v_j} \bh_{\theta} \big{(} \mathbf{u}, \mathbf{u}_{\mathbf{x}}, \mathbf{u}_{\mathbf{x} \mathbf{x} } \big{)}^{\top} \mathbf{ \wh{u}}  \bigg{]}  \\
&+ \sum_{j=1}^q  \frac{\partial^2}{\partial x_j^2} \bigg{[}  \nabla_{w_j} \bh_{\theta} \big{(} \mathbf{u}, \mathbf{u}_{\mathbf{x}}, \mathbf{u}_{\mathbf{x} \mathbf{x} } \big{)}^{\top}  \mathbf{ \wh{u} }\bigg{]}  \\
&+ \sum_{n=1}^{N_t-1} \delta( t- t_n) \nabla_{u} \norm{ \mathbf{u}(t_n, \bx; \nu_m) - \mathbf{V}(t_n, \bx; \nu_m) }, 
\label{DPMdetAdjoint_2}
\end{split}
\end{equation}
with the final condition
\begin{eqnarray}
\wh{\bu}(t_N, \mathbf{x}) = \nabla_{u} \norm{ \mathbf{u}(t_N, \mathbf{x}; \nu_m) - \mathbf{V}(t_N, \mathbf{x}; \nu_m) }, 
\end{eqnarray}
and the gradient of the objective function (\ref{ObjectiveFunction}) is then
\begin{eqnarray}
\nabla_{\theta}  L(\theta) = \int_0^{t_N} \int_{\Omega}  \wh{\bu}(t,\bx) \cdot \nabla_{\theta}  \bh_{\theta} \big{(} \bu(t,\bx), \bu_{\bx}(t,\bx), {\bu}_{\bx \bx}(t, \bx) \big{)}\, d \bx dt.
\end{eqnarray}

\subsection{Adjoint equations with a divergence-free constraint} \label{DiscreteAdjoint}

When subject to a divergence-free constraint $\nabla_{\mathbf{x}} \cdot \mathbf{u} = 0$, such as in the case of the incompressible Navier--Stokes equations augmented with a neural-network model $\bh_\theta$, the DPM framework (\ref{DPM}) becomes
%The incompressible Navier--Stokes equations include a divergence-free condition $\nabla_{\mathbf{x}} \cdot \mathbf{u}(t, \mathbf{x} ) = 0$ which is enforced by a pressure variable $p(t, \mathbf{x})$. The DPM framework (\ref{DPM}) can be extended to this case:
\begin{align}
\frac{\partial \mathbf{u} }{\partial t}   &= - \nabla_{\mathbf{x}} p +  \bbf_{\nu} \big{(} \mathbf{u}, \mathbf{u}_\mathbf{x}, \mathbf{u}_{\mathbf{x} \mathbf{x}} \big{)} + \bh_{\theta} \big{(} \mathbf{u}, \mathbf{u}_\mathbf{x}, \mathbf{u}_{\mathbf{x} \mathbf{x} } \big{)},  \\
\nabla_{\mathbf{x}} \cdot \mathbf{u} &= 0.
\label{DPM_df}
\end{align}
The divergence-free condition is enforced via the pressure-like variable $p$. Thus, for $x \in \Omega$, the pressure satisfies
\begin{eqnarray}
 \nabla^2 p  = \nabla_{\mathbf{x}} \cdot \bbf_{\nu} \big{(} \mathbf{u}, \mathbf{u}_\mathbf{x}, \mathbf{u}_{\mathbf{x} \mathbf{x}} \big{)} +  \nabla_{\mathbf{x}} \cdot  \bh_{\theta} \big{(} \mathbf{u}, \mathbf{u}_\mathbf{x}, \mathbf{u}_{\mathbf{x} \mathbf{x} } \big{)}.
\label{DPM_df_Pressure}
\end{eqnarray}

For computation, it is advantageous to time discretize (\ref{DPM_df}) via an operator splitting method and then derive the adjoint equations for the time-discretized equations. This ensures that the adjoint equations are fully compatible with the discretized PDE.  A standard projection method is used for this~\cite{Chorin1968}:
\begin{align}
\mathbf{u}^{\ast} (t + \Delta t, \mathbf{x})  &= \mathbf{u}(t, \mathbf{x}) + \Delta t\left[  f_{\nu} \big{(} \mathbf{u}(t,\mathbf{x}), \mathbf{u}_\mathbf{x}(t,\mathbf{x}), \mathbf{u}_{\mathbf{x} \mathbf{x}}(t, \mathbf{x}) \big{)} + h_{\theta} \big{(} \mathbf{u}(t, \mathbf{x}), \mathbf{u}_\mathbf{x}(t, \mathbf{x}), \mathbf{u}_{\mathbf{x} \mathbf{x} }(t, \mathbf{x} ) \big{)} \right], \\
\nabla_{\mathbf{x}}^2 p(t, \mathbf{x} ) &= \frac{ \nabla_{\mathbf{x}} \cdot \mathbf{u}^{\ast} (t + \Delta t, \mathbf{x})}{ \Delta t}, \\
\mathbf{u} (t + \Delta t, \mathbf{x})  &= \mathbf{u}^{\ast} (t + \Delta t , \mathbf{x})  - \Delta t\nabla_{\mathbf{x}} p(t, \mathbf{x} ),
\label{OperatorSplitting}
\end{align}
which yields the adjoint 
\begin{equation}
\begin{split}
\nabla_{\mathbf{x}}^2 \widehat{p}(t, \mathbf{x} ) &= - \frac{ \nabla_{\mathbf{x}} \cdot \mathbf{ \wh{u}} (t + \Delta t, \mathbf{x})}{ \Delta t},  \\
\mathbf{\wh{u}}^{\ast} (t + \Delta t, \mathbf{x})  &= \mathbf{\wh{u}} (t + \Delta t , \mathbf{x})  + \nabla_{\mathbf{x}} \widehat{p}(t, \mathbf{x} ) \Delta t,  \\
\mathbf{\wh{u}}(t , \mathbf{x}) &= \mathbf{\wh{u}}^{\ast} (t + \Delta t, \mathbf{x}) + \bigg{\{}   \nabla_{u} \bbf_{\nu_m} \big{(} \bu, \bu_{\bx}, \bu_{\bx \bx} \big{)}^{\top}  \wh{\bu}^{\ast}   \\
&- \sum_{j = 1}^q  \frac{\partial}{\partial x_j} \bigg{[}  \nabla_{v_j} \bbf_{\nu_m} \big{(} \mathbf{u}, \mathbf{u}_{\bx}, \mathbf{u}_{\mathbf{x} \mathbf{x} } \big{)}^{\top} \mathbf{\wh{u}}^{\ast}  \bigg{]}  \\
&+  \sum_{j=1}^q \frac{\partial^2}{\partial x_j^2} \bigg{[}  \nabla_{w_j} \bbf_{\nu_m} \big{(} \mathbf{u}, \mathbf{u}_{\mathbf{x}}, \mathbf{u}_{\bx \bx} \big{)}^{\top}  \mathbf{\wh{u}}^{\ast} \bigg{]}  \\
&+  \nabla_{u} \bh_{\theta} \big{(} \mathbf{u}, \mathbf{u}_{\mathbf{x}}, \mathbf{u}_{\bx \bx} \big{)}^{\top} \mathbf{ \wh{u}}^{\ast}  \\
&- \sum_{j=1}^q \frac{\partial}{\partial x_j} \bigg{[} \nabla_{v_j} \bh_{\theta} \big{(} \mathbf{u}, \mathbf{u}_{\mathbf{x}}, \mathbf{u}_{\mathbf{x} \mathbf{x} } \big{)}^{\top} \mathbf{ \wh{u}}^{\ast}  \bigg{]}  \\
&+ \sum_{j=1}^q  \frac{\partial^2}{\partial x_j^2} \bigg{[}  \nabla_{w_j} \bh_{\theta} \big{(} \mathbf{u}, \mathbf{u}_{\mathbf{x}}, \mathbf{u}_{\mathbf{x} \mathbf{x} } \big{)}^{\top}  \mathbf{ \wh{u} }^{\ast}  \bigg{]} \bigg{ \} }  \Delta t  \\
&+ \sum_{n=1}^{N_t-1} \mathbf{1}(t = t_n) \nabla_{u} \norm{ \mathbf{u}(t_n, \bx; \nu_m) - \mathbf{V}(t_n, \bx; \nu_m) }, 
\label{OperatorSplittingAdjoint}
\end{split}
\end{equation}
where $\mathbf{1}(t=t_n)$ is unity when $t=t_n$ and zero otherwise.
The starting condition for the backward-in-time adjoint solve is
\begin{eqnarray}
\wh{\bu}(t_N, \mathbf{x}) = \nabla_{u} \norm{ \mathbf{u}(t_N, \mathbf{x}; \nu_m) - \mathbf{V}(t_N, \mathbf{x}; \nu_m) }.
\end{eqnarray}
The gradient of the objective function (\ref{ObjectiveFunction}) is\todo{I'm leary of the $t+\Delta t$ argument of the adjoint solution in the integrand.  Is this really right? \textcolor{red}{JS: It seems Ok to me. We define the discretized forward equations in (2.13-2.15).}}\
\begin{eqnarray}
\nabla_{\theta}  L(\theta) =  \Delta t \sum_{t= 0, \Delta t, \ldots, t_{N_t-1}} \int_{\Omega}  \wh{\bu}^{\ast}(t + \Delta t,\bx) \cdot \nabla_{\theta}  \bh_{\theta} \big{(} \bu(t,\bx), \bu_{\bx}(t,\bx), {\bu}_{\bx \bx}(t, \bx) \big{)} \, d \bx. 
\end{eqnarray}
This adjoint equation is used in our demonstrations.

\section{Sub-grid-scale closure for incompressible turbulence} 
\label{SGS}

The momentum equation, when filtered for LES, contains an unclosed sub-grid-scale (SGS) term that has been the subject of extensive modeling efforts. Accounting for it, along with any numerical discretization errors, is the goal of the deep learning model in this application. The Navier--Stokes equations are introduced in Section~\ref{DNS}, followed by a discussion of filtering and the LES governing equations in Section~\ref{LES}.

\subsection{Navier--Stokes governing equations}\label{DNS}

Incompressible fluid flow is governed by the
Navier--Stokes equations, which comprise a momentum balance
\begin{equation}
  \pp{u_i}{t}  + \pp{u_iu_j}{x_j} = -\frac{1}{\rho}\pp{p}{x_i} +  \frac{\mu}{\rho}\pp{}{x_j}\left(\pp{u_i}{x_j}+\pp{u_j}{x_i} -
  \frac{2}{3}\pp{u_k}{x_k}\delta_{ij}\right), \label{eq.mom}
\end{equation}
and an incompressibility constraint
\begin{equation}
  \pp{u_k}{x_k} = 0, \label{eq.mass}
\end{equation}
where $\rho$ is the fluid density, $\mu$ is the dynamic viscosity, $u_i$ is the $i$\textsuperscript{th} velocity component,
$p$ is the pressure, and  $\delta_{ij}$ is the
Kronecker delta. Repeated indices 
imply summation. We take $\rho$ and $\mu$
to be constant.

For DNS, \eqref{eq.mom}
and \eqref{eq.mass} are discretized with sufficient resolution that all turbulence scales are accurately resolved.  This typically requires a mesh spacing comparable to the 
Kolmogorov scale to achieve mesh independence for typical statistical observables~\cite{Moin1998}. Evidence suggests that the DNS time step must be smaller than the Kolmogorov time scale in order to avoid spurious dissipation~\cite{Choi1994}, though in general the time step size requirements depend on the application and numerical methods~\cite{Moin1998}. For many applications, which have Reynolds numbers that lead to excessively small Kolmogorov scales, DNS is prohibitively expensive even for relatively simple flows~\cite{Moin1998}.

% LES formulation [JFM]
\subsection{Large-eddy simulation} \label{LES}

Using LES, computational expense is reduced by resolving only the largest turbulence scales.  This involves, implicitly or explicitly, a spatial filtering operation applied to the Navier--Stokes equations, which leads to terms that depend on unrepresented scales. Thus, the filtered equations are said to be unclosed.  Sub-grid-scale models are introduced to represent the effect of unrepresented scales in terms of the resolved scales, thereby closing the equations. In practice, the accuracy of LES calculations can be improved by better sub-grid-scale closures. 

A filtered quantity is denoted $\ol{\phi}$, which indicates
\begin{equation}
  \ol{\phi}(\mathbf{x},t) = \int_\Omega G(\mathbf{r},\mathbf{x})\phi(\mathbf{x}-\mathbf{r},t)\,\d\mathbf{r},
  \label{eq.filter}
\end{equation}
where $\mathbf{x},\mathbf{r}\in\Omega\subset\mathbb{R}^p$ and the filter kernel is $G(\mathbf{r},\mathbf{x})$. Common choices are box, Gaussian, and spectral cutoff filters. We choose the common box filter for simplicity and because it replicates the common practice of implicit filtering, in which the filtering operation~\eqref{eq.filter} is not explicitly performed so the LES mesh spacing itself truncates the small-scale features.  A box filter on a uniform grid with spacing $\ol\Delta$ has $G=1$ within a ${\bar\Delta}^3$ cube centered at $\mathbf{x}$~\cite{Clark1979}. Elsewhere, $G=0$. 
% The filtering operation~\eqref{eq.filter} for an inhomogeneous box filter becomes
% \begin{equation}
%   \ol{\phi}(\mathbf{x},t) = \frac{1}{\Delta^3}\int_{-\Delta/2}^{\Delta/2}\int_{-\Delta/2}^{\Delta/2}\int_{-\Delta/2}^{\Delta/2} \phi(x_1-r_1,x_2-r_2,x_3-r_3,t) \d r_1 \d r_2 \d r_3.
%   \label{eq.box_filter}
% \end{equation}

% We note that while the filter kernel is typically assumed to commute with
% differentiation, this assumption only holds in general for homogeneous filters on uniform grids~\cite{Ghosal1995}.

Governing equations are obtained by applying \eqref{eq.filter} to \eqref{eq.mom} and \eqref{eq.mass}, which yields
\begin{align}
  \frac{\partial \ol{u}_i}{\partial t} &= - \frac{\partial\ol{u}_i \ol{u}_j}{\partial x_j}
    - \frac{1}{\rho}\frac{\partial \ol{p}}{\partial x_i}
    + \frac{\mu}{\rho}\frac{\partial}{\partial x_j}\left(\pp{\ol{u}_i}{x_j} + \pp{\ol{u}_j}{x_i} - \frac{2}{3}\pp{\ol{u}_k}{x_k}\delta_{ij}\right)
    + \pp{\tau_{ij}^r}{x_j} \label{eq.LESmomentum} \\
  \frac{\partial \ol{u}_k}{\partial x_k}  &= 0, \label{eq.LESmass}
\end{align}
where the SGS stress tensor $\tau_{ij}^r = \ol{u_iu_j}-\ol{u}_i\ol{u}_j$ requires closure.

\begin{figure}
  \centering
  \subfloat[Filtered DNS]{
    \includegraphics[width=0.3\textwidth]{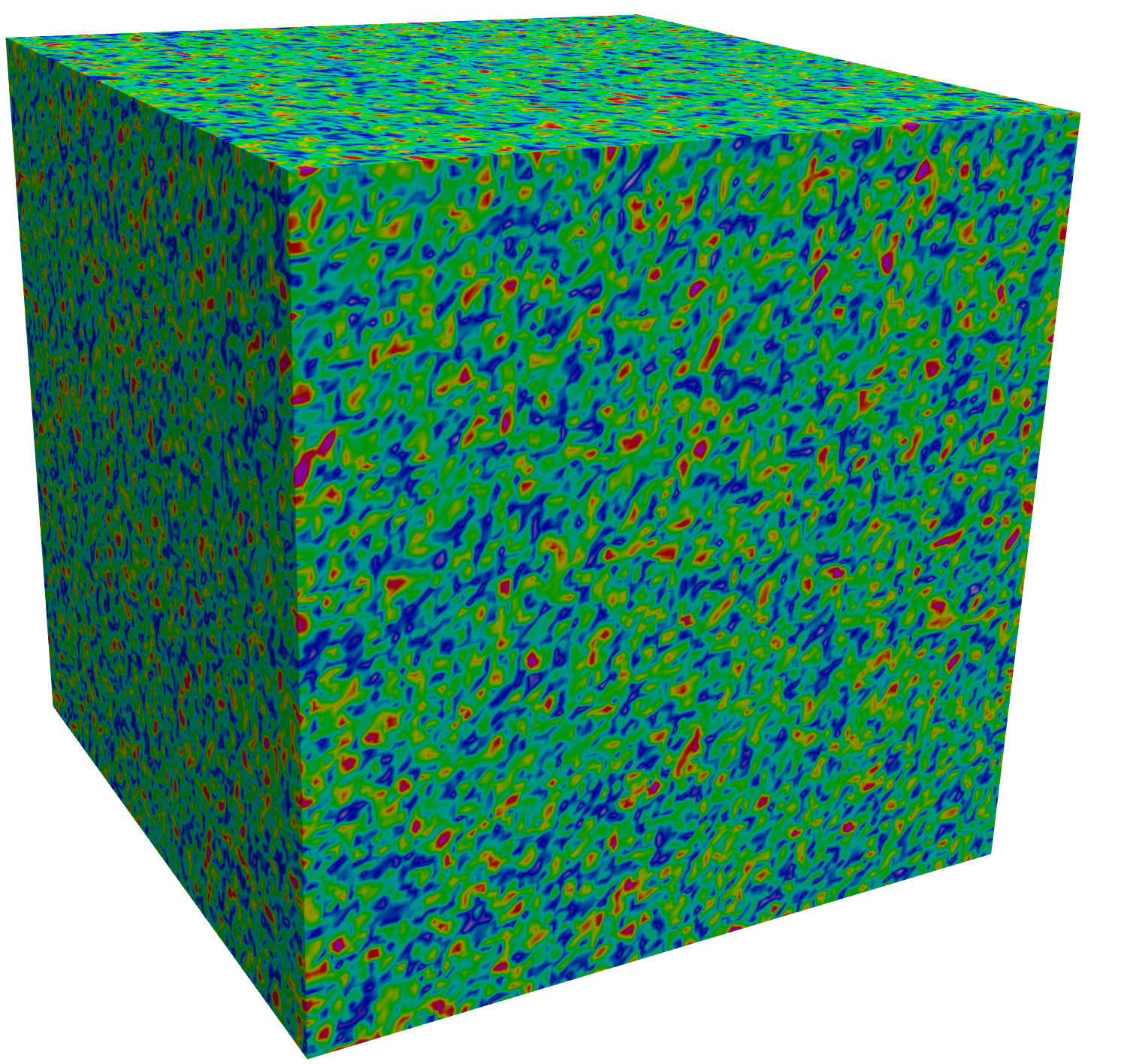}}
  \subfloat[LES, Dynamic Smagorinsky]{
    \includegraphics[width=0.3\textwidth]{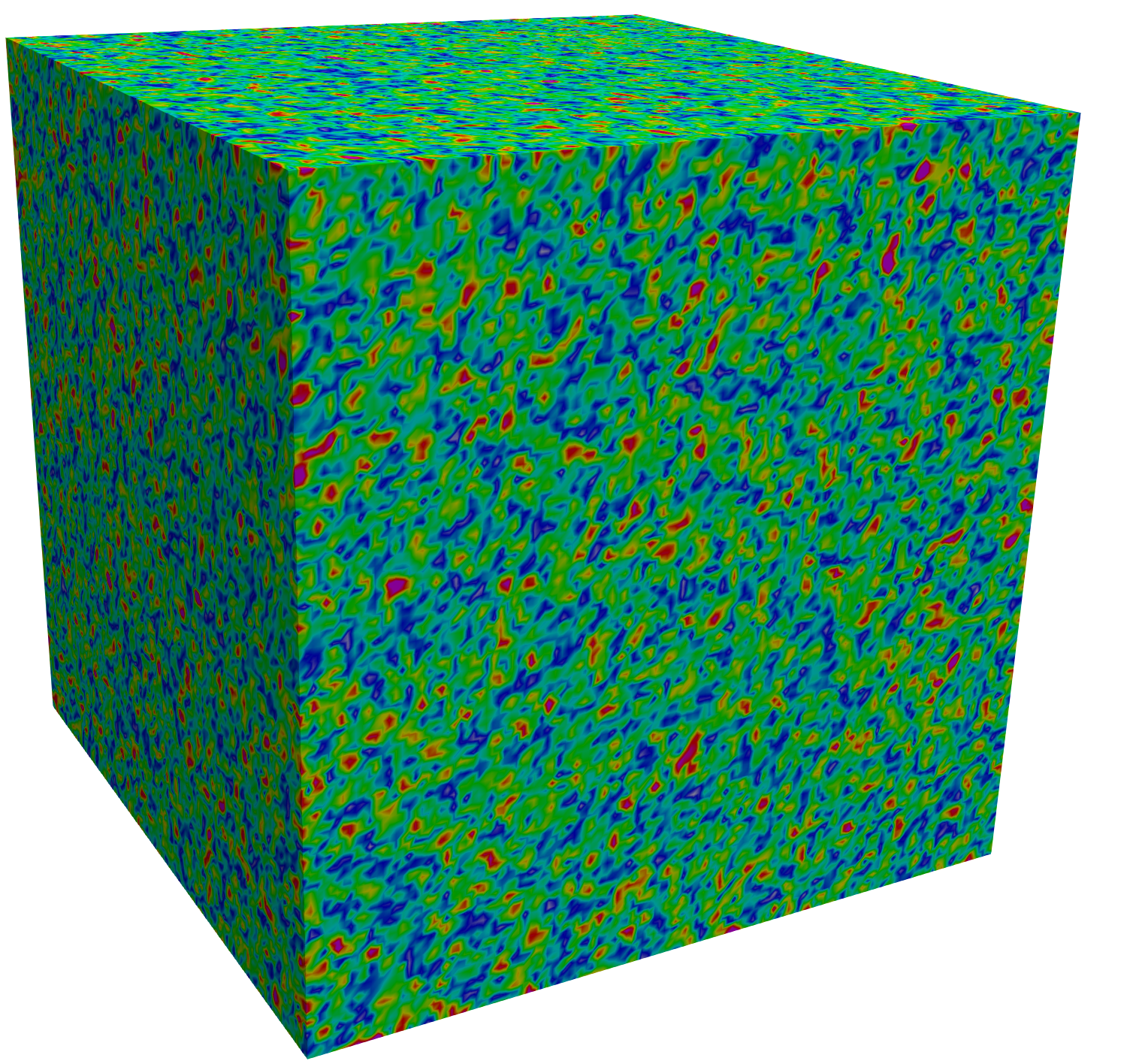}}
  %\subfloat[LES, Adjoint DPM]{
  %  \includegraphics[width=0.3\textwidth]{vort_surf_dnsbox_2048_Delta16_Down16_ML_crop.png}}
  \includegraphics[width=0.06\textwidth]{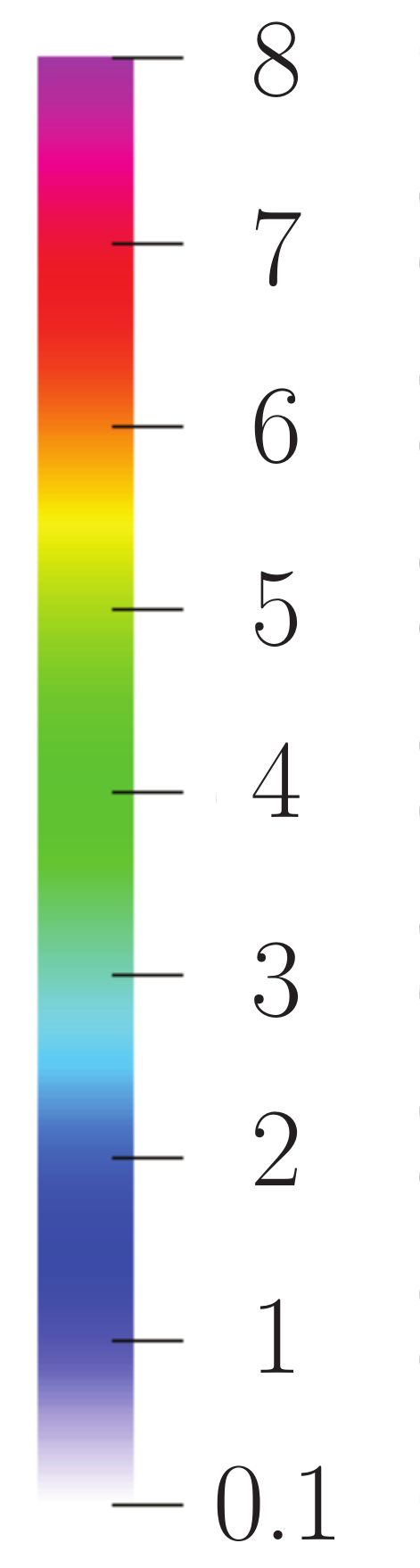}
  \caption{Vorticity magnitude ($\times 10$) in decaying isotropic turbulence, normalized by the eddy-turnover time scale $t_{\ell,0}=k/\varepsilon$  (see Section~\ref{DNS_data}), shown when energy is $36\,\%$ of the initial condition: (a) filtered $N=2048^3$ DNS using $\ol{\Delta}/\Delta x_\DNSsub=16$ for an effective grid resolution $N=128^3$, and (b) $N=128^3$ LES solution using the dynamic Smagorinsky~\cite{Germano1991} sub-grid-scale model.
}
  \label{fig.vorticity_2048}
\end{figure}

The most common closures are based on gradient-diffusion with the eddy-viscosity coefficient determined as suggested by Smagorinsky~\cite{Smagorinsky1963,Rogallo1984} or dynamically estimated based on resolved scales~\cite{Germano1991,Lilly1992}.  Other options include scale-similarity models~\cite{Bardina1980} and filter-scale Taylor expansions~\cite{Clark1979,Vreman1997,Vollant2016}. These and related models are extensively reviewed elsewhere~\cite{Rogallo1984,Lesieur1996,Meneveau2000}. Figure~\ref{fig.vorticity_2048} compares a $\ol{\Delta}/\Delta x=16$ filtered $N=2048^3$ DNS field on a $N=128^3$ grid with a corresponding $N=128^3$ LES using the dynamic Smagorinsky model. For this Reynolds number and relatively small LES filter width (\textit{i.e.}, a wide range of resolved scales), the dynamic Smagorinsky model yields a qualitatively-correct LES solution in Figure~\ref{fig.vorticity_2048}.  This is also supported quantitatively based on turbulence statistics such as we consider subsequently. 

\begin{figure}
  \centering
  \subfloat[Filtered DNS]{
    \includegraphics[width=0.3\textwidth]{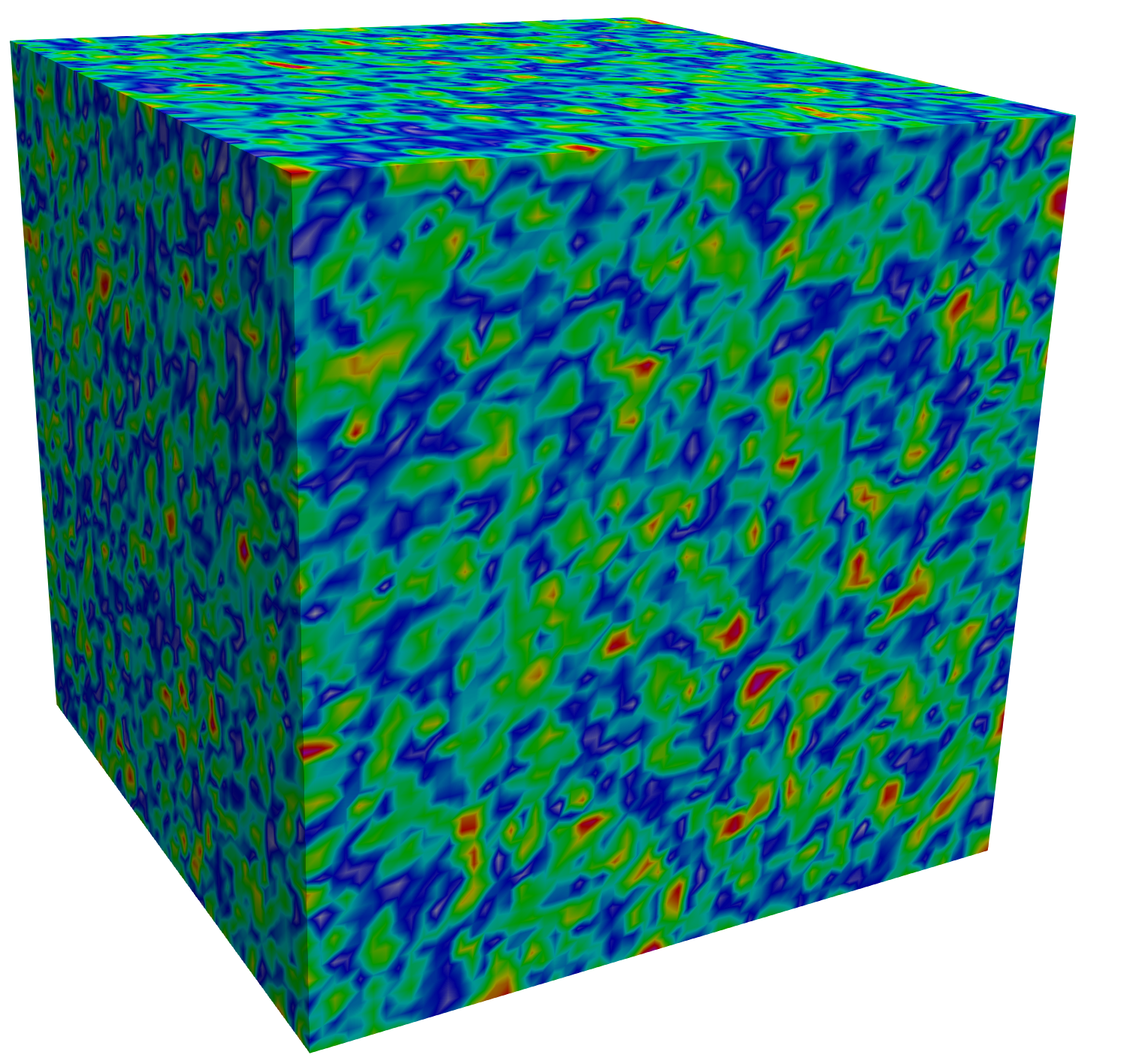}}
  \subfloat[LES, Dynamic Smagorinsky]{
    \includegraphics[width=0.3\textwidth]{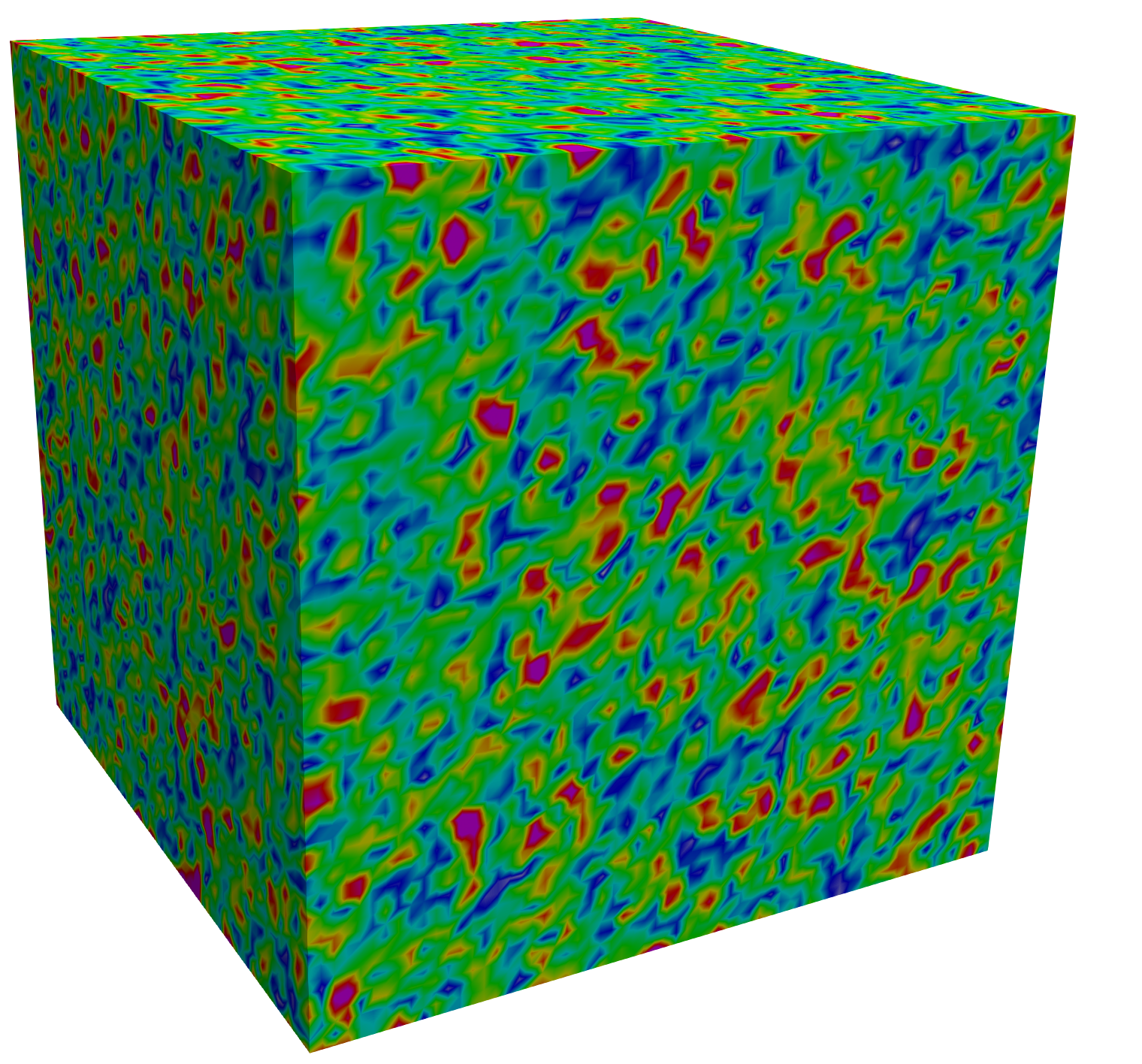}}
  \subfloat[LES, Adjoint DPM]{
    \includegraphics[width=0.3\textwidth]{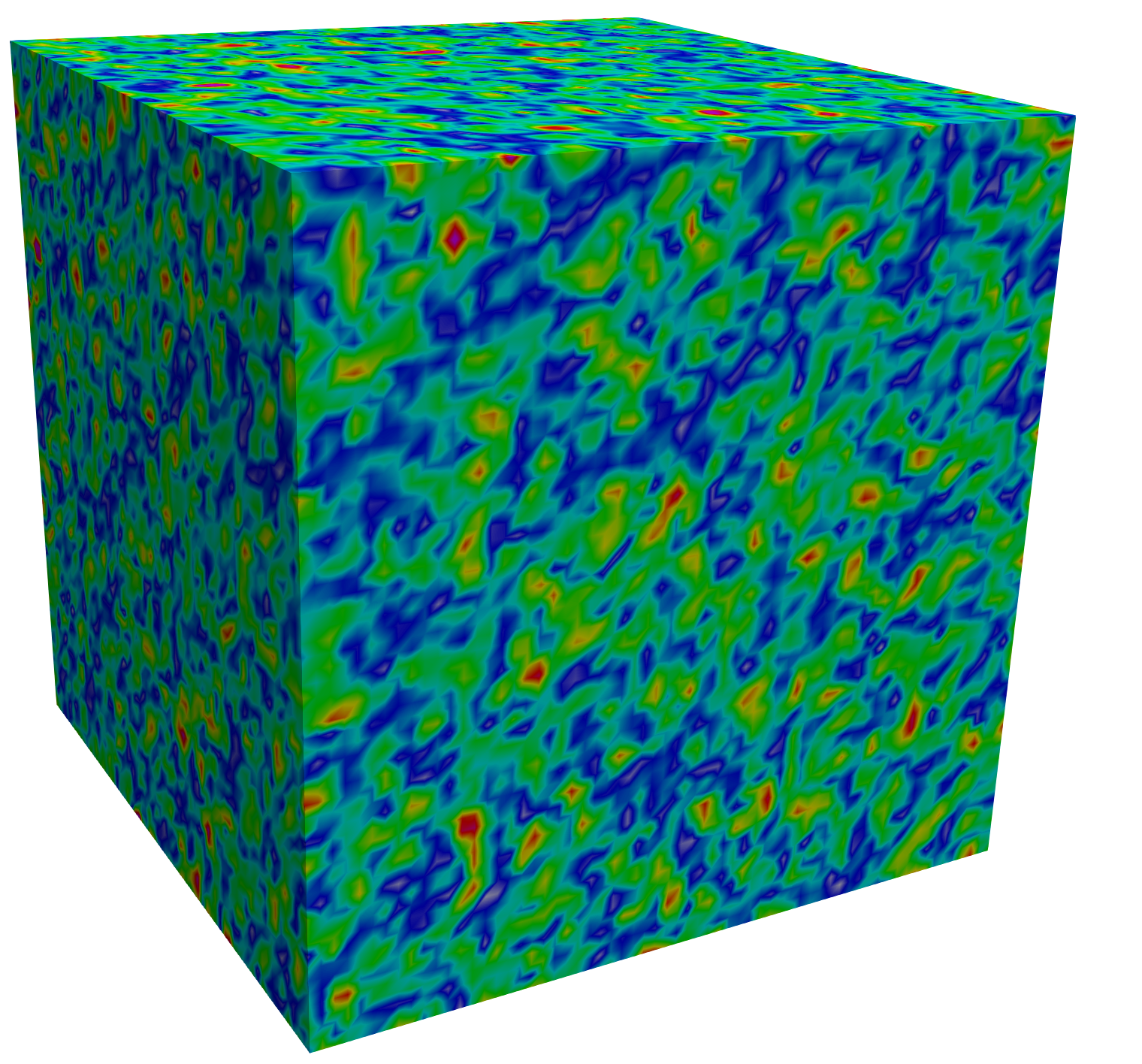}}
  \includegraphics[width=0.06\textwidth]{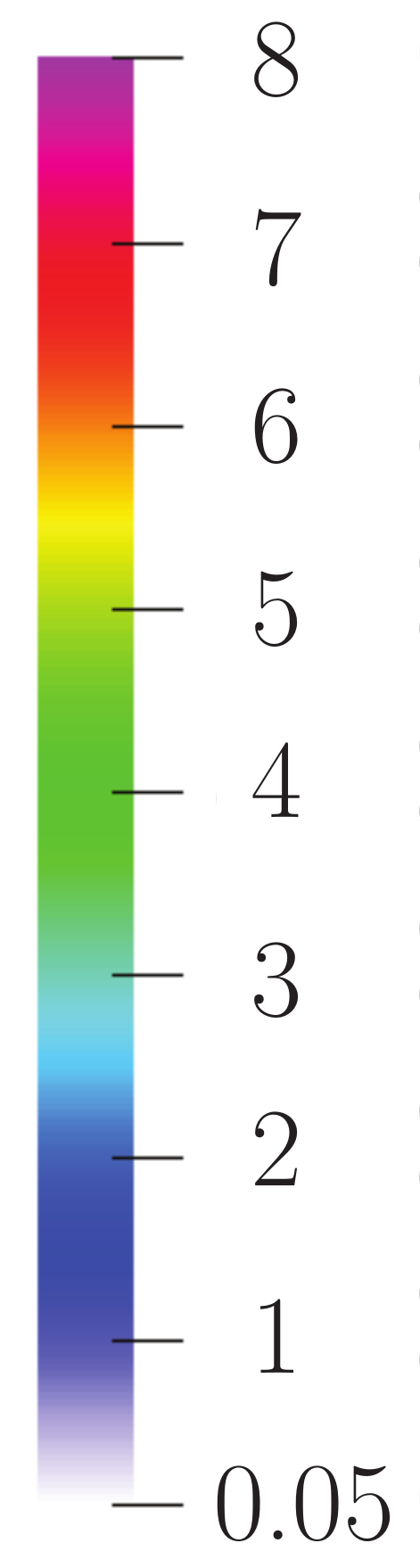}
  \caption{Vorticity magnitude ($\times 10$) in decaying isotropic turbulence, normalized by the eddy-turnover time scale $t_{\ell,0}=k/\varepsilon$ (see Section~\ref{DNS_data}), shown when energy is $9.2\,\%$ of the initial condition: (a) filtered $N=1024^3$ DNS using $\ol{\Delta}/\Delta x=16$ for an effective grid resolution $N=64^3$, (b)  $N=64^3$ LES solution using the dynamic Smagorinsky~\cite{Germano1991} sub-grid-scale model, and (c) $N=64^3$ LES solution using the adjoint-trained DPM model developed in this work.}
  \label{fig.vorticity_1024}
\end{figure}

While these common sub-grid-scale models are accurate for sufficiently small $\ol{\Delta}$ (and, if different, $\Delta x_\LESsub$), resolving still fewer scales would increase computational efficiency, if it can be done accurately. Even a modest improvement can significantly reduce the overall cost of a three-dimensional, time-dependent simulation.  Our objective is to develop turbulence closures using limited data available from such costly high-fidelity DNS data. Figure~\ref{fig.vorticity_1024}(a) shows a coarser  $\ol{\Delta}/\Delta x_\DNSsub=16$ filtered DNS solution on a $N=64^3$ grid, obtained from a $N=1024^3$ DNS. Figure~\ref{fig.vorticity_1024}(b) shows the analogous $N=64^3$ LES using the same dynamic Smagorinsky model. For this coarser filter (\textit{i.e.}, a narrower range of resolved scales), obvious visual differences suggest that the model produces a qualitatively-incorrect solution. Conversely, the adjoint-trained DPM model produces the qualitatively-correct solution shown in Figure~\ref{fig.vorticity_1024}(c), even for this coarse mesh. Section~\ref{Results} includes quantitative comparisons between the DPM and Smagorinsky and dynamic Smagorinsky models. 

\section{Discretization errors and learning corrections} \label{DiscreteUnknownPhysics}

%Our trusted data is a DNS dataset solved on a $N=1024^3$ grid. The DNS solution is filtered using the box filter~\eqref{eq.box_filter} and is restricted to coarser grid resolutions matching those used for LES:   $32^3$, $64^3$, and $128^3$.  These are used to provide a target for the LES model to match.

In addition to closure error, LES calculations suffer from discretization error. These discretization errors can themselves be viewed as unclosed terms, which arise from discretization of the governing equations on a coarse grid. 
Sub-grid-scale closures formulated as PDE terms, as well as any \textit{a priori}-trained machine learning turbulence models (see Section \ref{Apriori}), do not account for them. Yet in standard LES practice they can be comparable to $\btau_r$ modeling errors, depending on the filter width and spatial discretization scheme~\cite{Ghosal1996,Kravchenko1997,Chow2003}. This section discusses and quantifies discretization errors for LES and shows how the DPM can learn to compensate for discretization errors. This trained model will of course be linked to both the mesh density and the numerical schemes, and training dataset will need to be sufficiently large to allow for generalizations.  However, resolutions will not vary widely for efficient LES applications since it is most efficient to run with the coarsest mesh possible, which will aid generalization.  

The filtered Navier--Stokes solution $\ol{u}_i$ satisfies
\begin{align}
\frac{\partial \ol{u}_{i} }{\partial t} 
&= \mathcal{\chA}^{\Delta}_i( \ol{\bu}, \ol{p} ) 
+ \underbrace{\bigg{(}  \mathcal{A}_i( \ol{\bu} , \ol{p} )  
-  \mathcal{ \chA}_i^{\Delta} ( \ol{\bu}, \ol{p} ) \bigg{)}}_{\textrm{Discretization error}} + \underbrace{\phantom{\bigg(}\nabla\cdot\btau^r \phantom{\bigg)}}_{\text{Closure mismatch}}\notag \\
0 &=  \cD^{\Delta}(\ol{\bu} )  + \underbrace{\bigg{(} \nabla \cdot \ol{\bu} - \cD^{\Delta}(\ol{\bu} )  \bigg{)}}_{\textrm{Discretization error}},
\label{DiscreteEqnDecomposition}
\end{align}
where $\mathcal{A}_i ( \ol{\bu}, \ol{p}) = - \frac{\partial\ol{u}_i \ol{u}_j}{\partial x_j} - \frac{1}{\rho}\frac{\partial \ol{p}}{\partial x_i} + \frac{\mu}{\rho}\frac{\partial}{\partial x_j}\left(\pp{\ol{u}_i}{x_j} + \pp{\ol{u}_j}{x_i} - \frac{2}{3}\pp{\ol{u}_k}{x_k}\delta_{ij}\right)$ are terms in the continuous flow equation, $\mathcal{ \chA}_i^{\Delta} ( \ol{\bu}, \ol{p} )$ is the discrete approximation to $\mathcal{A}_i ( \ol{\bu}, \ol{p} )$ for the selected mesh and numerical schemes, and $\cD^{\Delta}(\ol{\bu} )$ is the corresponding discrete approximation to $ \nabla \cdot \ol{\bu}$.  When trained on the coarse LES mesh with the filtered (downsampled) DNS data, the DPM $\bh_{\theta}$ will learn the closure mismatch and discretization errors in (\ref{DiscreteEqnDecomposition}).

\begin{figure}
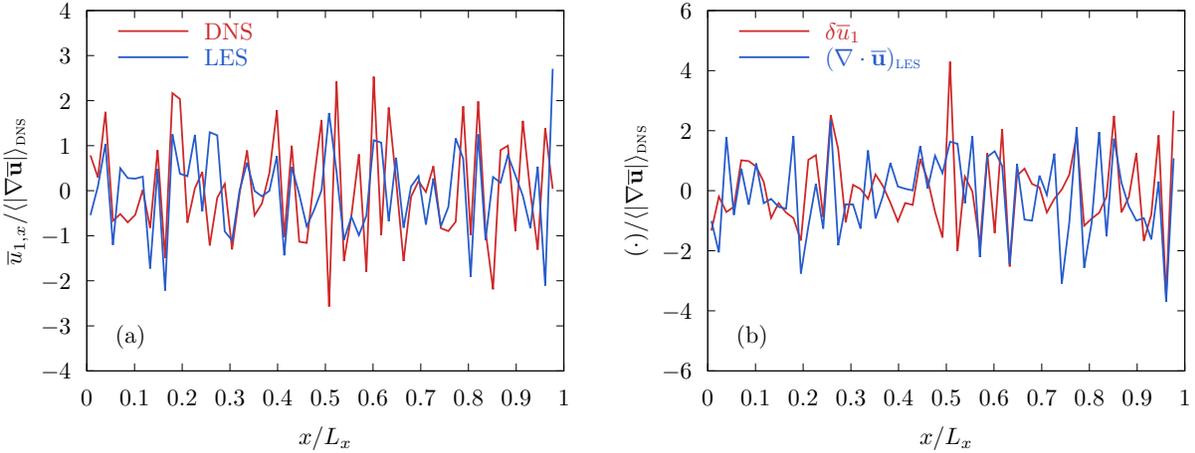

  \centering
  \subfloat{\resizebox{0.5\textwidth}{!}{\input{div_vel.tex}}}
  \subfloat{\resizebox{0.5\textwidth}{!}{\input{norm_error.tex}}}
  \caption{Typical errors versus position:  (a) finite-difference approximations to the filtered velocity gradient $\ol{u}_{1,x}$ evaluated on DNS ($N=2048^3$) and LES ($N=64^3$, $\ol{\Delta}/\Delta x=32$) grids, normalized by the average DNS velocity gradient magnitude; and (b) $\delta \ol{u}_1=\ol{u}_{1,x,\LESsub}-\ol{u}_{1,x,\DNSsub}$ and the normalized velocity divergence on the LES grid.}
  \label{FiniteDifferenceErrorFig}
\end{figure}

To assess the discretization errors in the LES, several quantities calculated with the finite-difference schemes used in the DNS and LES are listed in Table~\ref{TableFiniteDifferenceError}. 
% It shows the error in the finite-difference approximation to a filtered velocity gradient evaluated on the coarse LES grid, compared to the same filtered velocity gradient evaluated on the fine DNS grid. With $\Delta x=L_x/n_{x,\DNSsub}$ the DNS grid resolution and $\ol{\Delta}/\Delta x$ be the filter-width relative to the DNS grid space, and $n_{x,\DNSsub}/n_{x,\LESsub}$ be the LES grid coarsening factor. The equations are said to be implicitly filtered when $\ol{\Delta}/\Delta x=n_{x,\DNSsub}/n_{x,\LESsub}$.
\begin{table}
  \centering
\scalebox{.95}{
 \begin{tabular}{ c c c c c c c c }
   \toprule
Filtering &   $\frac{\ol{\Delta}}{\Delta x_\DNSsub}$ &  $\frac{\Delta}{\Delta x_\DNSsub}$ &
   $\frac{\langle|\delta \ol{u}_1|\rangle}{\langle|\nabla\ol{\mathbf{u}}|\rangle_\DNSsub}$ &
   $\mathrm{max}(\cD^{\Delta}(\ol{\bu}))_\DNSsub$ &
   $\mathrm{max}(\cD^{\Delta}(\ol{\bu}))_\LESsub$ &
   $\frac{\mathrm{max}(\cD^{\Delta}(\ol{\bu}))_\LESsub}{\langle|\nabla\ol{\mathbf{u}}|\rangle_\DNSsub}$ &
   $\frac{\langle|\cD^{\Delta}(\ol{\bu})|\rangle_\LESsub}{\langle|\nabla\ol{\mathbf{u}}|\rangle_\DNSsub}$ \\
   \midrule
\multirow{3}{*}{Implicit}   &      8 &      8  &  0.601 &  $2.545\times10^{-8}$ &  $2.257\times10^5$ &  7.893 & 0.827 \\
    &   16 &      16 &  0.854 &  $1.566\times10^{-8}$ &  $1.216\times10^5$ &  6.288 & 1.081 \\
        & 32 &      32 &  1.076 &  $0.596\times10^{-8}$ &  $0.579\times10^5$ &  6.651 & 1.208 \\
        \midrule
\multirow{3}{*}{Explicit}   &        32 &      16 &  0.654 &  $6.399\times10^{-9}$ &  $4.734\times10^4$ &  5.435 & 0.895 \\
    &        32 &      8  &  0.325 &  $7.195\times10^{-9}$ &  $2.855\times10^4$ &  3.277 & 0.476 \\
       & 32 &      4  &  0.141 &  $6.930\times10^{-9}$ &  $1.325\times10^4$ &  1.522 & 0.211  \\
   \bottomrule
\end{tabular} }
  \caption{Error in the finite-difference approximation to the filtered velocity gradient $\delta\ol{u}_1=\ol{u}_{1,x,\LESsub}-\ol{u}_{1,x,\DNSsub}$, evaluated on LES grids of varying resolution $n_{x,\DNSsub}/n_{x,\LESsub}$, relative to the average magnitude of the filtered velocity gradient evaluated on the DNS grid.  The maximum of the discrete filtered-velocity divergence is also shown evaluated on the DNS and LES grids.  Averages $\langle \cdot \rangle$ are over the full simulation domain at fixed time.}
\label{TableFiniteDifferenceError}
\end{table}
The starting-point DNS velocity field ${u}_i$ is computed on a $N=2048^3$ mesh. Filtered velocity components $\ol{u}_i$ are obtained with a box filter of width $\ol{\Delta}$.  With the DNS providing the trusted solution, we take as reference the $\ol{u}_i$ derivative in the $x$-direction at $(t,x+\Delta x/2)$ to be a dense-mesh finite difference
\begin{equation}
  \ol{u}_{i, x,\DNSsub}(t,x+\Delta_\LESsub/2) = \frac{ \ol{u}_i(t, x + \Delta x_\DNSsub ) -  \ol{u}_i(t, x) }{ \Delta x_\DNSsub},
  \label{DNSestimateOfDerivative}
\end{equation}
which is compatible with how the derivatives were calculated in the DNS.  On the coarse LES mesh, the corresponding filtered velocity gradient is evaluated as
\begin{equation*}
  \ol{u}_{i, x,\LESsub} (t,x+\Delta_\LESsub/2) = \frac{ \ol{u}_i(t, x + \Delta_\LESsub) -  \ol{u}_i(t, x )   }{ \Delta_\LESsub}  ,
\end{equation*} 
where $\Delta_\LESsub$ is the coarse-mesh resolution. Table~\ref{TableFiniteDifferenceError} shows that the difference between the DNS-mesh derivative and the LES-mesh derivative $\delta \ol{u}_i=\ol{u}_{i,x,\LESsub}-\ol{u}_{i,x,\DNSsub}$ is comparable to the average velocity gradient magnitude evaluated on the DNS mesh $\langle|\nabla\ol{\mathbf{u}}|\rangle_\DNSsub$. Figure~\ref{FiniteDifferenceErrorFig}(a) compares them for a representative segment of the domain.  In Table~\ref{TableFiniteDifferenceError}, the error $\delta \ol{u}_1$ increases with the mesh size, becoming greater than the average velocity gradient magnitude for $\ol{\Delta}/\Delta x_\DNSsub=32$. With such errors, even if the unclosed term $\nabla\cdot\btau^r$ is modeled exactly, the LES calculation would be inaccurate.

It is noteworthy that the discrete divergence-free constraint $\cD^{\Delta}(\ol{\bu} )= 0$  is not satisfied by the filtered DNS solution when evaluated on the coarse mesh. Table~\ref{TableFiniteDifferenceError} shows that the filtered velocity satisfies the discrete divergence-free condition on the DNS grid within a small factor of the Poisson equation solver tolerance ($10^{-9}$), while the filtered velocity on the coarse LES grid does not. $\cD^{\Delta}(\ol{\bu} )$ is comparable to the average velocity gradient, as is shown in Figure~\ref{FiniteDifferenceErrorFig}(b). Similarly, the average divergence residual is comparable to the average DNS velocity gradient magnitude.

Numerical errors such as these are widely recognized. Nonetheless, LES solvers typically enforce the condition $\cD^{\Delta}(\ol{\bu} ) = 0$.  This accurately reflects the commutative property of the linear divergence operator~\eqref{eq.mass} and filter~\eqref{eq.filter}, though it does not reflect the discretization error of the discrete divergence operator evaluated on the coarse LES grid. Enforcing $\cD^{\Delta}(\ol{\bu} ) = 0$ to hold exactly for the coarse LES grid is a choice.  (Perot~\cite{Perot2011} discusses this more generally for incompressible flow simulation: the common practice of enforcing a discrete-exact version of incompressibility rather than a discrete-exact version of momentum is also a choice.)  Another perspective is that the coarse-mesh (LES) divergence should not be zero in the LES.  If there were a hypothetical sub-grid-scale field (and not necessarily a unique one~\cite{Langford1999}) that exactly reflected the realistic SGS dynamics, the coarse-mesh sampling of it would not yield $\cD^{\Delta}(\ol{\bu} ) = 0$, and would deviate further from this condition the coarser the mesh. In an extension in Section~\ref{sec.model_non_div_free},  the DPM is shown to also be able to exploit this flexibility to provide an accurate model that reproduces key features of the turbulence by learning, rather than explicitly representing, the physics of the incompressibility constraint.

\section{Numerical results} \label{Results}

The DNS data used for training and out-of-sample testing are introduced in Sections~\ref{DNS_data} and~\ref{PPM}, the neural network architecture is described in Section~\ref{Hyperparameters}, and model comparisons are presented in Section~\ref{sec.model_comparison}. Their behavior is compared to a more straightforward \textit{a priori} training regimen in Section~\ref{Apriori}. The extension mentioned at the end of the previous section, which learns rather than enforces strict adherence to the divergence-free $\cD^\Delta(\ol{\bu}) = 0$ constraint, is discussed in Section~\ref{sec.model_non_div_free}. Finally, computational cost is quantified and discussed in Section~\ref{ComputationalCost}.

\subsection{DNS data} \label{DNS_data}

The decaying isotropic turbulence DNS were initialized with a standard spectrum~\cite{Passot1987} at an initial Reynolds number  $Re_{t,0} = \rho u_{\text{rms}}\ell/\mu=1749$, where $u_{\text{rms}} = \langle u_i u_i \rangle^{1/2}$ is the domain-averaged root-mean-squared velocity, $\ell=k^{3/2}/\varepsilon$ is the pseudo-integral scale of the initial spectrum, $k=\langle u_iu_i\rangle/2$ is the turbulence kinetic energy (TKE), and $\varepsilon$ is the TKE dissipation rate. The reference density and viscosity are $\rho=1$ and $\mu_0=1$.

Though isotropic turbulence is nominally Reynolds-number isoparametric, we chose to train the DPM in a manner more generally applicable.  We require that the model, if possible, learn the Reynolds number scaling. Though obvious for isotropic turbulence, this is emblematic of requiring the model to learn corresponding parameters in more complex scenarios.  For training and testing cases, we adjust the dimensional time scales of the turbulence such that the model inputs vary by factors of 10 for nominally the same instantaneous Reynolds number in the decay.  However, this is a side concern:  the main point is that the DPM learns the instantaneous structure of the turbulence sufficiently to close the LES governing equations (including numerical error terms) better than established models.

Table~\ref{tbl.DNS_IC} lists initial parameters for the six $N=1024^3$ DNS datasets that are used to train and test models. Three each are used for training and out-of-sample testing. The decay time scales are adjusted via $\mu$ and $u_{\text{rms},0}$ for fixed initial Reynolds number $Re_{t,0}$. Data were produced for $\mu/\mu_0=\{0.5,0.75,1.0,1.25,1.5,2.0\}$, each for two random phasings of the initial conditions, for a total of 12 simulations.  As shown in Table~\ref{tbl.DNS_IC}, the initial TKE dissipation rate varies by a factor of 64.
\begin{table}
  \centering
  \begin{tabular}{c c c c c c c c c c}
    \toprule
    $N$ & Case & {$\mu/\mu_0$} & {$u_{\text{rms},0}$} & {$t_{\ell,0} = k/\eps$} & {$t_{\eta,0} = (\mu\eps/\rho)^{1/2}$} &
        {$\eps_0$} & Train & Test & {$Re_{t,0}$} \\
    \midrule
    \multirow{6}{*}{$1024^3$}
    & 1 & 0.50 & 101 & $1.27\times10^{-1}$ & $2.03\times10^{-3}$ & $1.21\times10^5$ & $\bullet$ & & \multirow{6}{*}{1749}\\
    & 2 & 0.75 & 152 & $8.49\times10^{-2}$ & $1.35\times10^{-3}$ & $4.09\times10^{5}$ & & $\bullet$ & \\
    & 3 & 1.00 & 203 & $6.38\times10^{-2}$ & $1.01\times10^{-3}$ & $9.69\times10^{5}$ & $\bullet$ & & \\
    & 4 & 1.25 & 255 & $5.09\times10^{-2}$ & $8.12\times10^{-4}$ & $1.89\times10^{6}$ & & $\bullet$ & \\
    & 5 & 1.50 & 304 & $4.24\times10^{-2}$ & $6.79\times10^{-4}$ & $3.27\times10^{6}$ & & $\bullet$ & \\
    & 6 & 2.00 & 406 & $3.18\times10^{-2}$ & $5.09\times10^{-4}$ & $7.74\times10^{6}$ & $\bullet$ & & \\
    \bottomrule
  \end{tabular}
  \caption{The DNS datasets that are used for training and testing the models are described. All cases have an initial integral scale $L_{ii}=1.00$. The domain for all cases is a periodic cube with sides of length $L=66.5$, the initial Kolmogorov length scale for all cases is $\eta=3.18\times10^{-2}$, and the initial Reynolds number based on the Taylor microscale is $Re_\lambda\approx 162$. Cases used for training and testing ensembles are indicated.}
  \label{tbl.DNS_IC}
\end{table}

% Dimensional quantities used in simulations, for reference
%  rho = 1.2 kg/m^3
%  mu  = 1.8678e-5 Pa*s
%  eta = 2.15e-5 m
%  Physical domain size L=4.5 cm
%  Initial integral length scale $L_{ii}=6.77\times10^{-4}$\,m.
%
% Normalization factors
%  L_norm = L_ii = 6.77e-4 m (gives L/L_norm = 66.47)
%  u_norm = 2.299e-2 m/s
%  t_norm = 2.945e-2 s

The DNS were first allowed to decay for $t \approx 0.05 t_{\ell,0}$. After this period, the velocity fields were stored every $0.01 t_{\ell,0}$. These were box filtered with $\ol{\Delta}/\Delta x_\DNSsub=16$. Initial conditions on the coarse LES mesh were obtained by restricting data from the fine DNS mesh, resulting in the irrecoverable information loss inherent to LES~\cite{Vreman1997,Langford1999}. For simplicity, we only consider cases in which the filter width and LES grid resolution are equivalent, which is common practice though it lacks significant theoretical basis~\cite{Lund2003,Bose2010}.

These DNS have a large number of integral scales per domain length, $L/L_{ii}\approx 66$, which is helpful for statistical sampling. For comparison, recent simulations aimed at maximizing the Reynolds number have $L/L_{ii}\approx5$~\cite{Ishihara2016}, which is nearly correlated across the periodic domain.
  
The DNS datasets were produced using the \textit{NGA} code~\cite{Desjardins2008,MacArt2016}, which uses a fractional-step method~\cite{Kim1985}. For the DNS cases listed in Table~\ref{tbl.DNS_IC}, the initial CFL number was 0.4, and the time step $\Delta t$ was fixed throughout the simulations. Space is discretized using second-order central differences on a staggered mesh~\cite{Harlow1965}.

% . The \textit{NGA} source code has been extensively optimized for large-scale simulations. Approximately 1.3~million core-hours were expended to generate 300\,TB of DNS data.  It is well-understood that a Fourier spectral method would, in some sense, be optimal for this configuration.  However, its parallel scaling is challenging, and a key strength of the DPM is how it copes with truncation errors, as discussed in Section~\ref{DiscreteUnknownPhysics}.  The finite-difference algorithm illustrates this property.

Adjoint-based \textit{a posteriori} model training was performed using a new implementation that matches the \textit{NGA} finite-difference discretization but is Python-native and interfaces with the \textit{PyTorch}~\cite{PyTorchNIPS} machine learning and automatic differentiation library. It was co-verified with \textit{NGA} and validated against standard accepted turbulence results. While the adjoint could in theory be entirely computed using existing automatic differentiation software, this would be prohibitively expensive in practice. Instead, we solve the adjoint equations \eqref{OperatorSplittingAdjoint} on the staggered mesh using the Python-native solver. Only gradients of the neural network ($\nabla_{\theta} \mathbf{h}_{\theta}$) are evaluated using automatic differentiation.

% Our code is accelerated by  accelerated by  and solve it for $\widehat{\bu}$ as described in Section~\ref{Hyperparameters}.

\subsection{Pre-processing of training data} \label{PPM}

It is important to clearly define the training problem, including how the training data is processed, which specifies the unclosed terms targeted by $\bh_\theta$.  The following develops the specifics for our demonstration, building upon the general discussion of discretization errors in Section~\ref{DiscreteUnknownPhysics}.

\begin{table}
\begin{center}
\begin{tabular}{ccl}\hline
Field & Mesh $\Delta x$ & Description \\\hline
$\bu$ & DNS & trusted DNS data\\
$\fbu$ & DNS & filtered DNS data (\ref{eq.filter}) \\
$\fbU$ & LES & sub-sampled filtered DNS data (every $k$-th point for $k = \frac{\Delta x_\DNSsub}{\Delta x_\LESsub}$) \\
$\fbw$ & LES & divergence-free projected filtered DNS training target\\
$\hbu$ & LES & numerical solution of LES equations \\\hline
\end{tabular}
\caption{Notation for data variables used in simulations and model training.}\label{tab:vars}
\end{center}
\end{table}

Variable labels associated with different stages are summarized in Table~\ref{tab:vars}.  We assume that the DNS fields $\bu$ are sufficiently trusted that we disregard any errors they might entail.
The filtered DNS solution $\fbu$ notionally  satisfies the continuous filtered flow equations,
\begin{equation}
\begin{split}
\frac{\partial \ol{u}_{i} }{\partial t} &=   \mathcal{ A}_i( \ol{\bu} , \ol{p} ) + \nabla\cdot\btau^r  \\
0 &=  \nabla \cdot \ol{\bu},
\end{split}
\label{RecallLES}
\end{equation}
where we use the compact $\cA$ notation of (\ref{DiscreteEqnDecomposition}) for the terms in the flow equations. This stage only includes the explicit sub-grid-scale stress closure $\nabla\cdot\btau^r$, and it remains represented numerically on the fine $\Delta x_\DNSsub$ mesh.  To be used in training on the $\Delta x_\LESsub$ coarse mesh, 
it is downsampled:
$\fbU_{ijk} = \fbu(t, i \Delta_\LESsub, j \Delta_\LESsub, k \Delta_\LESsub)$, where for simplicity $\fDelta = \Delta x_\LESsub$.  The downsampled filtered DNS satisfies
\begin{equation}
\begin{split}
\frac{\partial \ol{U}_{ijk} }{\partial t} &= \mathcal{\chA}^{\Delta}( \ol{\mathbf{U}}, \ol{p} )_{ijk} +\big{[}  \mathcal{ A}( \ol{\bu} , \ol{p} )(i \fDelta, j \fDelta, k \fDelta) - \mathcal{\chA}^{\Delta}( \ol{\mathbf{U}}, \ol{p} )_{ijk} \big{]} + \nabla\cdot\btau^r(i \fDelta, j \fDelta, k \fDelta)\\
0 &= \cD^{\Delta}(\ol{ \mathbf{U}} )_{ijk} + \big{[} \nabla \cdot \ol{\bu}(i \Delta, j \Delta, k \Delta) - \cD^{\Delta}(\ol{ \mathbf{U} } )_{ijk} \big{]}, \label{Uevolution}
\end{split}
\end{equation}
where on this coarse mesh $\fbU$ is not divergence free due to the residual $\nabla \cdot \ol{\bu}(i \Delta, j \Delta, k \Delta) - \cD^{\Delta}(\ol{ \mathbf{U} } )_{i,j,k}$.
Appendix~\ref{a:discreteforms} includes details on the discrete forms underlying the projection used to ensure that the training data $\fbw$ satisfies $\cD^\Delta(\fbw) = 0$ in order to be compatible with standard LES practice (though this is revisited in Section~\ref{sec.model_non_div_free}).  The divergence-free result of this projection can be expressed $\mathbf{w}= \ol{ \mathbf{U} } + \mathcal{H}^{\Delta} \ol{ \mathbf{U} },$ where $\mathcal{H}^{\Delta}$ is a (constant) linear operator with the property that $\norm{ \mathcal{H}^{\Delta} \ol{ \mathbf{U} }_t } \rightarrow 0$ as $\Delta \rightarrow 0$.  This final projection stage adds still further unclosed terms to the momentum equation, now governing $\fbw$ on the coarse LES mesh: 
\begin{equation}
\begin{split}
\frac{\partial \mathbf{\bw}_{ijk} }{\partial t} &=  \mathcal{\chA}^{\Delta}(  \mathbf{\bw} , \ol{p} )_{ijk} +  \underbrace{\phantom{\bigg[} \nabla\cdot\btau^r\phantom{\bigg]}}_{\text{SGS closure}}    
+ \underbrace{ \bigg{[}   \mathcal{\chA}^{\Delta}( \ol{\mathbf{U}}, \ol{p} )_{ijk} -  \mathcal{\chA}^{\Delta}( \mathbf{\bw} , \ol{p} )_{ijk}  \bigg{]} }_{\textrm{Projection effect on $\cA^\Delta$ evaluation}} \\
&+  \underbrace{\frac{\partial \mathcal{H}^{\Delta} \ol{\mathbf{U}} }{\partial t}}_{\textrm{Change in the projection}}
+ \underbrace{ \bigg{[}  \mathcal{ A}( \ol{\bu} , \ol{p} )(i \Delta, j \Delta, k \Delta)  -   \mathcal{\chA}^{\Delta}( \ol{\mathbf{U}}, \ol{p} )_{ijk} \bigg{]} }_{\textrm{Finite-difference error}}
\label{ProjectedLES}
\end{split}
\end{equation}
with the discrete divergence-free constraint
\begin{eqnarray}
\cD^{\Delta}(\mathbf{w} )_{ijk} =  0.
\end{eqnarray}
The last four terms of (\ref{ProjectedLES}) are, in a sense, unclosed, representing the mismatch targeted by the neural network $\bh_{\theta}$ by optimizing $\theta$ in 
\begin{equation}
\frac{\partial\hbu}{\partial t} = \chA^\Delta(\hbu,\ol{p}) + \bh_\theta,
\end{equation}
where $\hbu$ is a numerical LES solution, and optimization seeks to minimize 
\begin{equation}
L(\theta) =  \sum_{n=1}^{N_t} \sum_{i, j, k = 1}^N \norm{\hbu - \fbw}.
\end{equation}

\subsection{Model architecture and hyperparameters} \label{Hyperparameters}

The deep neural network $F_{\theta}(z)$ has the following architecture:
\begin{equation}
\begin{split}
  H^1 &= \sigma(W^1 z + b^1)   \\
  H^2 &= \sigma(W^2 H^1 + b^2)    \\
  H^3 &=  G^1 \odot H^2 \qquad\qquad \text{with} \quad  G^1 = \sigma(W^5 z + b^5) \\
  H^4 &= \sigma(  W^3 \odot H^3 + b^3)  \\
  H^5 &=   G^2 \odot H^4 \qquad\qquad \text{with} \quad G^2 = \sigma(W^6 z + b^6) \\
  F_{\theta}(z) &= W^4 H^5  + b^4 
\end{split}\label{eq.NN}
\end{equation}
where $\sigma$ is a $\tanh()$ element-wise nonlinearity, $\odot$ denotes element-wise multiplication, and the parameters are $\theta = \{ W^1, W^2, W^3, W^4, W^5, W^6, b^1, b^2, b^3, b^4, b^5, b^6 \}$. The final machine learning model $\bh_{\theta}$, which is used in the DPM (\ref{DPM}), applies a series of derivative operations on $F_{\theta}(z)$. The input $z$ to $F_{\theta}(z)$ at an LES grid point $(i,j,k)$ includes the velocity components and their first and unmixed second derivatives at $(i,j,k)$ as well as the 6 closest neighboring grid points. This selection does not strictly enforce Galilean invariance, but this was not found to be a challenge. The issue of Galilean invariance should be considered further, especially in regard to further extrapolation from the training data that we consider here.  Enforcing this (or many other) invariances would be straightforward.  Each layer includes $N_\textsc{H} = 200$ hidden units, for a total of approximately $245,000$ parameters, all initialized using a standard Xavier initialization~\cite{Xavier}. Inputs to the neural network are normalized by the same set of constants for all cases.  Although large, this network proves both effective (Section~\ref{sec.model_comparison}) and efficient (Section~\ref{ComputationalCost}).

Training is distributed across multiple compute nodes, with each working with a randomly-selected velocity field from the training data in Table~\ref{tbl.DNS_IC}. We advance the LES solutions over five LES time steps, which is the equivalent of 50 DNS time steps, then solve the adjoint over the equivalent reverse-time interval.  
Parameter updates use the RMS\-prop algorithm with a standard decaying learning rate magnitude schedule. Training is accelerated by distributing computations across multiple GPU nodes.

% -------------------------------------------------------
% Results for divergence-free models

\subsection{Out-of-sample model comparison} \label{sec.model_comparison}

\begin{figure}
  \centering
  \resizebox{0.65\textwidth}{!}{\input{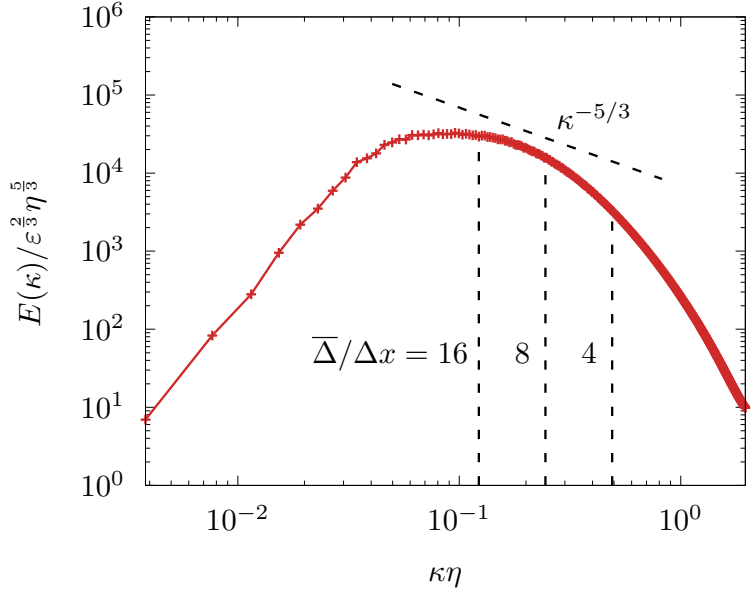}}
  \caption{Energy spectrum of the unfiltered $N=1024^3$ isotropic turbulence DNS. The nominal cutoff wavenumbers for spectrally-sharp filters of size $\ol{\Delta}/\Delta x=4$, $8$, and $16$ are indicated by vertical dashed lines. A reference Kolmogorov $\kappa^{-5/3}$ inertial range is also indicated.}
  \label{fig.spectrum_DNS}
\end{figure}

To quantify model needs, and ultimately performance, in terms of representing turbulence scales, velocity fields are Fourier transformed in all three directions to provide  standard wavenumber-magnitude $\kappa\equiv(\kappa_i\kappa_i)^{1/2}$ spectra $E(\kappa)$.
Figure~\ref{fig.spectrum_DNS} shows this for the full range of $\kappa$ for the unfiltered $N=1024^3$ DNS data with respect to the nominal filter cutoffs for different cases.  The most challenging $\ol{\Delta}/\Delta x_\DNSsub=16$ filter nominal cutoff scale is close to the most energetic turbulence scales.  The $\ol{\Delta}/\Delta x_\DNSsub=4$ and $\ol{\Delta}/\Delta x_\DNSsub=8$ cutoffs, more typical for LES, include more represented scales and should therefore be more accurate.  However, the higher resolution they afford is costly:  halving the filter increases the operation count by approximately a factor of 16. Models that perform well for large $\ol{\Delta}$ are therefore attractive.

\begin{figure}
  \centering
  \resizebox{0.65\textwidth}{!}{\input{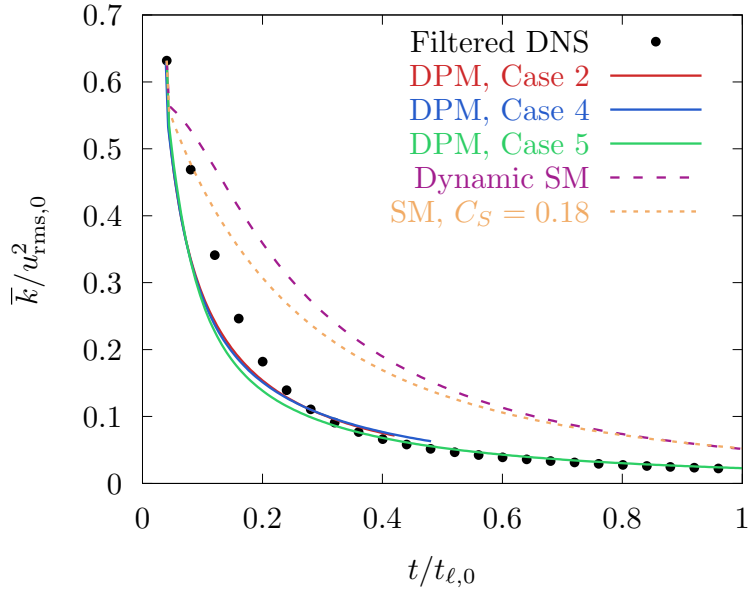}}
  \caption{Decay of $\ol{k}(t)$ for out-of-sample test case~2 ($\mu/\mu_0=0.75$), case~4 ($\mu/\mu_0=1.25$), and case~5 ($\mu/\mu_0=1.5$) from Table~\ref{tbl.DNS_IC}, for LES filter size $\ol{\Delta}/\Delta x=16$. The adjoint PDE-trained DPM correctly learns the normalized evolution despite not having been trained on these dissipation rates. }
  % divergence-free test/train configuration
  \label{fig.test_decay_divFree}
\end{figure}

The simplest quantity of interest for isotropic turbulence is the decaying kinetic energy of the resolved scales, $\ol{k}(t) \equiv\frac{1}{2}\langle\ol{u_iu_i}-\ol{u}_i\ol{u}_i\rangle$. Figure~\ref{fig.test_decay_divFree} compares $\ol{k}(t)$ with the filtered DNS.  For all the three out-of-sample test cases listed in Table~\ref{tbl.DNS_IC}, the DPM outperforms the widely-used Smagorinsky model~\cite{Smagorinsky1963,Rogallo1984}, whether the coefficient is determined dynamically~\cite{Germano1991,Lilly1992} or fixed at $C_S=0.18$~\cite{Clark1979}.  These  all have the same Reynolds number and so follow the same decay-rate profile, though this was learned only based on the instantaneous fields at different initial dissipation rates, so the DPM indeed learns the correct rescaling.  The eddy viscosity-type LES models perform better (not shown) with smaller $\ol{\Delta}$ (and thus significantly higher cost) but perform poorly for $\ol{\Delta}/\Delta x_\DNSsub = 16$.

\begin{figure}
  \centering
%  \subfloat{
   \resizebox{0.7\textwidth}{!}{\input{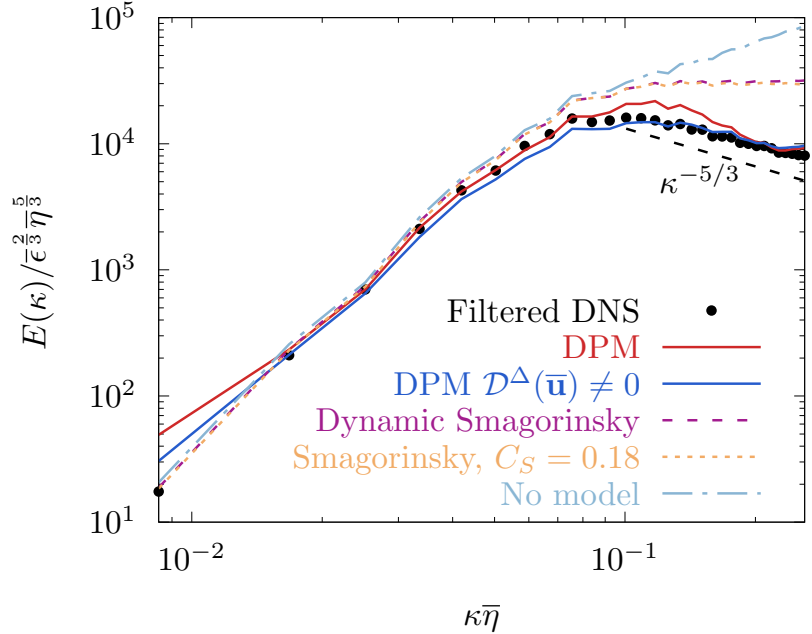}}
%}
%  \subfloat{ %\resizebox{0.5\textwidth}{!}{\input{Bij_ALL_dnsbox_1024_Lx0_045_NR_Delta16_Down16_visc1_5_00000100.tex}}}
  \caption{Resolved energy spectra % and (b) scaled velocity double correlations
    for filter size $\ol{\Delta}/\Delta x=16$ for an out-of-sample ($\mu/\mu_0=1.5$) test case at time instant $t/t_{\ell,0}=0.36$. DPM ($\cD^\Delta(\ol{\bu})\neq 0$) shows the non-divergence-free model result of Section~\ref{sec.model_non_div_free}.
}
  \label{fig.spectra_vel_corr_LES_2}
\end{figure}

The more detailed comparison in Figure~\ref{fig.spectra_vel_corr_LES_2} shows that the adjoint-trained DPM reproduces the spectrum remarkably well, qualitatively better than the eddy-viscosity-based models.  For this large $\ol{\Delta}/\Delta x_\DNSsub=16$, the constant-coefficient and dynamic Smagorinsky energy spectra are nearly identical, though it is confirmed that, as expected, the dynamic model outperforms the constant-coefficient model for smaller $\ol{\Delta}$. The overpredicted high-wavenumber energy is consistent with the underpredicted energy decay rates in Figure~\ref{fig.test_decay_divFree}. Also shown is a no-model LES, which is still further from the filtered DNS spectrum. 
% The spatial separation $r$ is normalized by the Taylor microscale $\lambda$, and the double correlations are normalized by the rms velocity. All normalization constants are evaluated from the filtered DNS data. 

\subsection{Comparison with \emph{a priori} training } \label{Apriori}

For comparison, we analyze the common practice of \emph{a priori} training, which seeks sub-grid-scale corrections without regard to how they would couple with the LES PDEs. To do this, $\btau^r$ from the filtered DNS data is used to directly train $\bh_{\theta}$ by minimizing  
\begin{equation}
  J(\theta) = \norm{ \bh_{\theta}(\ol{\bu} ) - \nabla\cdot\btau^r },
  \label{AprioriObjFunction}
\end{equation}
where $\ol{\bu}$ and $\btau^r$ are from the DNS without additional processing and do not change during the optimization.
This de-coupled form also does not include errors due to the LES discretization. Once trained, using the same input and neural-network architecture as the DPM models introduced in Section~\ref{Hyperparameters}, $\bh_{\theta}$ closes the LES equation.

An advantage of \textit{a priori} training is that it is easy to implement and only requires standard machine learning tools to minimize $J$. However, the DPM is expected to outperform \emph{a priori} training because of the mathematical inconsistency of the latter:  the \textit{a priori} approach interchanges optimization (training) with a nonlinear operation (the LES equations), which is expected to degrade subsequent predictions. For our demonstration of  \textit{a priori} training, $\bh_{\theta}$ is trained on the trusted filtered DNS data $\ol{\bu}$,  whereas during \emph{a posteriori} LES it only receives input from the solution of the discretized LES equations.   This provides robustness to nonlinear accumulation of finite-difference error during the simulation. In summary, the neural network parameters $\theta$ that minimize (\ref{AprioriObjFunction}) for the DNS data do not necessarily minimize the objective function (\ref{ObjectiveFunction}), and so do not necessarily yield a consistent and accurate LES model.

\begin{figure}
  \centering
  \resizebox{0.5\textwidth}{!}{\input{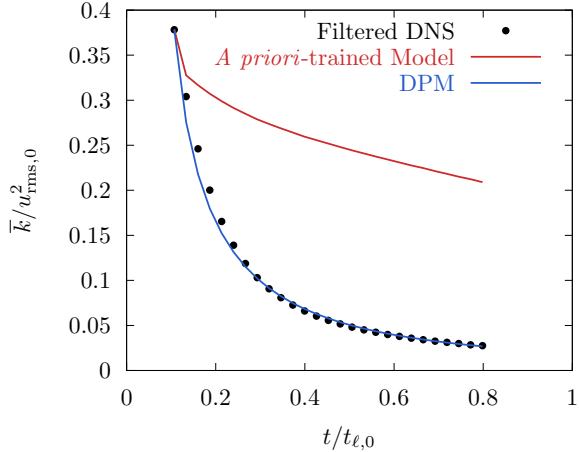}}
  \caption{Performance of an \emph{a priori}-trained deep learning model compared to an adjoint-trained DPM and filtered DNS data for filter size $\ol{\Delta}/\Delta x=16$ and case $\mu/\mu_0 = 1.0$ from Table~\ref{tbl.DNS_IC}.}
  \label{AprioriFig1}
\end{figure}

Figure~\ref{AprioriFig1} shows the poor predictive performance of the \emph{a priori}-trained deep learning closure model for LES. The DPM, trained with the adjoint method, performs substantially better than the \emph{a priori}-trained model. 

In summary, there are three main advantages sought with our adjoint-based approach.  Foremost, it avoids the inconsistency of \emph{a priori} training: optimization does not commute with a nonlinear function, which introduces error. 
Second, it accounts for numerical errors on the coarse LES grid. Including the full numerical discretization of the PDE in the adjoint-based optimization allows $\bh_\theta$ to respond to the unavoidable LES discretization error (discussed in Section~\ref{DiscreteUnknownPhysics}). This is linked to the particular resolution, but is better than neglecting this important challenge inherent in any LES.  Finally, \emph{a priori} training as designed in this example requires a full training-target description of the to-be-modeled $\nabla\cdot\btau_r$.  This is readily available in DNS, though not generally.  For example, \textit{a priori} training cannot so readily be used to estimate a deep learning PDE model from limited experimental data, which is almost always relatively sparse. It is mostly readily applied when the unclosed terms to be modeled are directly available as in (\ref{AprioriObjFunction}).  If the DPM is estimated from $M_s$ high-fidelity numerical simulation datasets and $M_e$ experimental datasets with $N_e$ measurement points, our original objective function~\eqref{ObjectiveFunction} becomes
\begin{equation}
\begin{split}
L(\theta) &= \underbrace{ \sum_{ \phantom{..} m = 1, \ldots, M_s}  \sum_{n=1}^{N_t} \int_{\Omega} \norm{ \mathbf{u}(t_n, \mathbf{x}; \nu_m) - \mathbf{V}(t_n, \mathbf{x}; \nu_m) }\, d \mathbf{x}}_{\textrm{Simulations}} \\&+ \underbrace{ \sum_{\ \phantom{..} m = M_s+1, \ldots, M_s+M_e}  \quad\sum_{n=1}^{N_t} \sum_{j =1}^{N_e} \norm{ \mathbf{u}(t_n, \mathbf{x}_j; \nu_m) - \mathbf{V}(t_n, \mathbf{x}_j; \nu_m) }}_{\textrm{Experiments}}. %\notag \\
\end{split}
\end{equation}
where $\bx_j$ and $t_n$ in the second sum do not need to be full-field quantities.  The DPM facilitates evaluations for whatever limited data points are available.

\subsection{Relaxation of the divergence-free constraint} \label{sec.model_non_div_free}

As discussed in Section~\ref{DiscreteUnknownPhysics}, the divergence-free constraint in the LES equations~\eqref{eq.LESmass} is not satisfied by filtered DNS data on the LES mesh. The standard approach, followed thus far, is to enforce incompressibility on the filtered target data as discussed in Section~\ref{PPM}. This ensured consistency between the discretized LES and the filtered target data:  the target data is a solution of the LES equations on the discrete grid.

We consider an alternate approach in this section. Instead of projecting the target data to be divergence-free, we relax the strict requirement that the LES solution be discretely divergence-free. This significantly reduces computational cost by eliminating the elliptic solve in the pressure-projection part of the algorithm. Of course, the filtered DNS data still satisfies the divergence-free constraint~\eqref{eq.mass}  on the DNS grid, so the model learns the divergence-free constraint from the evolution of the full-resolution DNS data rather than relying on its discrete enforcement on the LES grid.  As such, this is also a demonstration of the DPM's ability to learn physics that is omitted from the PDE.

A spectrum for this DPM ($\cD^\Delta(\ol{\bu})\ne 0$) case is also plotted in Figure~\ref{fig.spectra_vel_corr_LES_2}. Performance of this new model is comparable to the original DPM, better in some sense, particularly in matching low-wavenumber energy. However, the differences between the two deep learning models are small, particularly when compared to the improvement of both over the established LES models. The possibility of extending the non-divergence-free approach to more complex flows, in particular pressure-driven flows and reacting flows, is not automatically precluded.  
This example was included primarily as a further illustration of the capacity of the DPM framework to replace physics with learned models and closures.  We do recognize that less explicitly represented physics is likely to diminish the capacity for extrapolation, and that some form of validation would be needed as for any model reduction, machine learning or otherwise.  Our expectation, in this case, is some lost capacity to extrapolate to flows that include an important non-zero mean pressure gradient.

\subsection{Computational cost} \label{ComputationalCost}

LES is used to reduce cost, so we provide a brief analysis of this.  A general concern with deep learning models is high expense due to the many tensor operations to evaluate a large network.  For the cases reported thus far, our model has $248{,}418$ parameters, which includes a correspondingly large number of operations at each grid point each time step compared to the other models. Even in this case, the DPM, though based on a large network, is not uncompetitive.  We address this empirically for the current implementation in two parts. First, we quantify the cost of evaluating the neural network---and the associated solution accuracy---as a function of the number of parameters. Second, for reference, we compare the measured full-simulation cost of our Python-native LES to that of the optimized \textit{NGA} (Fortran) solver.

%, referencing the respective accuracy of the methods.

%Although the ultimate goal is to enable the learning of predictive models when there are not currently models available, even in this case the DPM, though based on a large network, is not uncompetitive.

Table~\ref{TableComputationalCostNeuralNetwork} lists the computational time required to evaluate the deep neural network (\ref{eq.NN}) at all points in a $N=64^3$ mesh for different network sizes. This accounts for almost 50\,\% of the total LES cost (see Table~\ref{tbl.cost_single}).  Although the operation count is high, the cost of this evaluation benefits tremendously from the highly-optimized (\textit{PyTorch}-based) implementation.  Table~\ref{TableComputationalCostNeuralNetwork} provides a measure of this by comparing the network-evaluation cost for a single CPU core (AMD 6276 Interlagos) and a single GPU accelerator (NVIDIA K20X).  The efficiency of the GPU implementation of the neural network evaluation enables bringing seemingly large trained networks into current practice for prediction.  Although cost increases with network size, the GPU-accelerated speedup also improves for the relevant range considered here.
%and characteristics of any such network evaluations that make them well-suited to GPU architectures
\begin{table}
  \centering
  \begin{tabular}{ c c c c c}
   \toprule
   {$N_\textsc{H}$}  & {$d_\theta$} &  {Time, CPU (s)} & Time, GPU (s) & {Speed-up}    \\
   \midrule
   5   &  4{,}278   & 1.83   & $4.98 \times 10^{-2}$ & 36.7 \\
   25  &  22{,}318  & 2.45   & $7.00 \times 10^{-2}$ & 35.0 \\
   50  &  47{,}118  & 3.35   & $7.75 \times 10^{-2}$ & 43.2 \\
   100 &  104{,}218 & 5.59   & $8.87 \times 10^{-2}$ & 63.0 \\
   200 &  248{,}418 & 10.49  & $1.53 \times 10^{-1}$ & 68.5 \\
   \bottomrule
 \end{tabular} 
 \caption{Average time to evaluate the deep neural network (\ref{eq.NN}) on a $N=64^3$ mesh. The network size is given in terms of the number of hidden units per layer $N_\textsc{H}$ and the total number of parameters $d_\theta$. The average wall-clock time is reported for a single AMD 6276 Interlagos CPU core and a single NVIDIA K20X GPU. The listed speed-up is the ratio of single-core CPU time to GPU time.}
\label{TableComputationalCostNeuralNetwork}
\end{table}

\begin{figure}
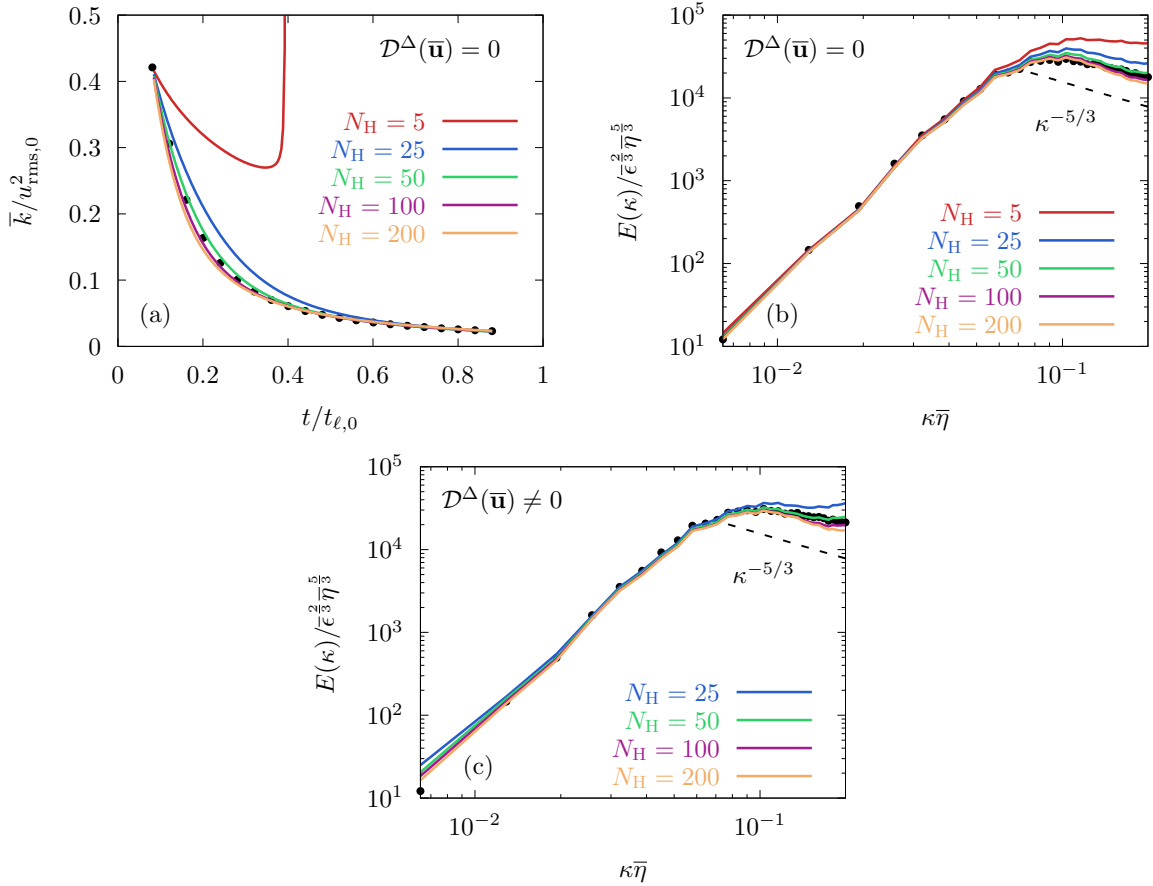

  \centering
  \resizebox{0.48\textwidth}{!}{\input{k_decay_H_projected.tex}}
  \resizebox{0.48\textwidth}{!}{\input{spectrum_H_dnsbox_1024_Lx0_045_NR_Delta16_Down16_visc1_5_50.tex}} \\
  \resizebox{0.48\textwidth}{!}{\input{spectrum_H_NDF_dnsbox_1024_Lx0_045_NR_Delta16_Down16_visc1_5_50.tex}}
  
 \caption{Dependence on number of hidden units per layer $N_\textsc{H}$ (lines) compared to filtered DNS (points):  (a) decay of $\ol{k}(t)$ and (b) spectra for divergence-free ($\cD^\Delta(\bu)=0$) models, and (c) spectra for $\mathcal{D}^\Delta(\bu)\ne0$ models. The filtered DNS fields in (a) and (b) are projected onto divergence-free manifolds. Spectra in (a) and (b) are at $t/t_{\ell,0}=0.2$. In (c), the $N_\textsc{H}=5$ case has become unstable at this time, and the $N_\textsc{H}=25$ is showing signs of high-wavenumber divergence. All models are trained on $\mu/\mu_0 \in\{0.5,1.0,2.0\}$ and are tested on $\mu/\mu_0=1.5$.}
   
   \todo[inline]{The $N_H = 200$ case in (b) does not seem to agree with that in Figure 6.  I see a different at the lowest wavenumbers.}
   \todo[inline]{The models were trained differently, as described in the legend.}

  \label{DPM_Hcomparison}
\end{figure}

The large $N_\textsc{H} = 200$ network we have considered was selected to focus our study on the DPM rather than the neural-network design.  Although it is able to leverage GPU acceleration to be practical at this large scale, smaller networks of similar design perform nearly as well, as shown in Figure~\ref{DPM_Hcomparison}.  This compares DPM $\ol{k}$ decay and spectra for $N_\textsc{H}\in\{5,25,50,100,200\}$ for both the divergence free $\cD^\Delta(\ol{\bu})=0$ and not strictly divergence-free $\cD^\Delta(\ol{\bu})\ne0$ variants.  Even $N_\textsc{H} = 25$ show better agreement with spectra than the Smagorinsky models (Figure~\ref{fig.spectra_vel_corr_LES_2}).  More hidden units (larger $N_\textsc{H}$) increase accuracy, with the associated cost.  We do not attempt to prove stability of the DPM models, which would likely be challenging.  No stabilizing limiters were used in the present implementation, though if needed they could provide a practical means of ensuring stability, such as are typically used to stabilize dynamic Smagorinsky models~\cite{Vreman1997}.  For both $\mathcal{D^\Delta}(\bu)=0$ and $\mathcal{D^\Delta}(\bu)\ne0$ variants, deeper neural networks reduce numerical instability, with $N_\textsc{H}\ge 50$ required for long-time ($t \geq 10^{-3}$) stability for $\mathcal{D^\Delta}(\bu)=0$ and $N_\textsc{H}\ge 100$ without this constraint ($\mathcal{D^\Delta}(\bu)\ne0$).

Of course, significant inefficiency of the flow solver within the overall DPM model would also hide the network evaluation cost.  To assess this, we compare the Python-based, \textit{PyTorch}-coupled LES solver used against the optimized Fortran \textit{NGA} solver.  These data, computed by inserting system timers around the time-marching loop and advancing the solution 50 steps, are shown in Table~\ref{tbl.cost_single}. The Python solver is less than a factor of two slower.
\begin{table}
  \centering
 \begin{tabular}{ c c c }
   \toprule
   Model  &  \multicolumn{1}{c}{Fortran (s/step)} & \multicolumn{1}{c}{Python (s/step)} \\
   \midrule
   No-model LES & 7.57 & 14.11  \\
   Smagorinsky, $C_S=0.18$  & 7.73 & 14.24  \\
   Smagorinsky, Dynamic & 8.44 & --- \\
   DPM ($N_\textsc{H} = 200$) & --- & 27.86  \\
   \bottomrule
 \end{tabular} 
  \caption{Single-core CPU performance of Fortran and Python LES solvers for decaying isotropic turbulence on a $N=64^3$ mesh. All computations were performed on a single AMD 6276 Interlagos CPU core. The dynamic Smagorinsky model was not implemented in Python, and the DPM was not implemented in Fortran.}
\label{tbl.cost_single}
\end{table} 

Finally, we provide an estimate of application computational cost for comparable solution accuracies. We use the \textit{NGA} Smagorinsky model on a $N=128^3$ mesh to produce a comparably accurate solution to the DPM on a $N=64^3$ mesh. Both solvers use the available computational resources on a Cray XK7 compute node containing 16 CPU cores and one NVIDIA K20X GPU. The Fortran solver is faster in single-core performance but, like many legacy solvers, is not readily GPU-accelerated. The network evaluation in the Python-native solver is easily GPU-accelerated using the \textit{PyTorch} library.  Performance is compared in Table~\ref{tbl.cost_multicore}. The Python-native DPM solution, even in divergence-free form, is approximately one-quarter the cost. Invoking the ability of the DPM to learn the divergence-free condition reduces the DPM-solution cost to approximately one-twentieth of the legacy-solution cost. 
\begin{table}
  \centering
 \begin{tabular}{ c c c }
   \toprule
   \multirow{2}{*}{Model}& \multicolumn{1}{c}{Fortran (s/step)} & \multicolumn{1}{c}{Python (s/step)} \\
   &  {$N=128^3$, 16-core} & {$N=64^3$, K20X GPU} \\
   \midrule
   %No-model LES & 6.42 & 1.31  \\
   Smagorinsky, $C_S=0.18$  & 6.47 & --- \\ %& 1.32  \\
   Smagorinsky, Dynamic & 6.73 & --- \\
   DPM & --- & 1.60  \\
   DPM ($\cD^\Delta(\ol{\bu}) \ne 0$) & --- & 0.31    \\
   \bottomrule
 \end{tabular} 
 \caption{Single-node performance of Fortran and Python LES solvers for comparable solution accuracy. The Smagorinsky models are evaluated on a $N=128^3$ mesh using the Fortran solver, and the deep learning models are evaluated on a $N=64^3$ mesh using the Python solver. The Fortran solver used 16 CPU cores on a Cray XK7 compute node (AMD 6276 Interlagos CPUs; 313~GF peak performance), while the Python solver used an NVIDIA K20X GPU (1.31~TF peak performance).}
\label{tbl.cost_multicore}
\end{table}

\section{Conclusion} \label{FutureWork}

Results demonstrate the promise of deep learning PDE models (DPMs) applied to problems with unresolved physics, demonstrating it on a case with limited high-fidelity numerical data for training. As formulated, however, the estimation method can also incorporate experimental data.  It was demonstrated for learning the unclosed terms in the filtered Navier--Stokes equations (the LES sub-grid-scale stress tensor). The DPM is trained on filtered DNS data and its out-of-sample accuracy is studied. 

The DPM recovered an accurate representation of the resolved turbulence.  This result is an important first step in demonstrating the ability of the model to learn unrepresented (or unknown) physics. It outperforms established eddy-viscosity models, including the dynamic Smagorinsky model, in terms of reproducing the resolved kinetic energy decay rate and the resolved kinetic energy spectrum observed in the exact filtered DNS data.  It is expected that additional optimizations and refinement of the design of the training regimen would further increase performance.
Results also suggest that the adjoint PDE-based training of the DPM is able to correct for numerical discretization errors, which are widely known to affect the performance of LES calculations on coarse meshes. These discretization errors were interpreted as additional unclosed terms in the discrete governing equations on the coarse LES grid.  

As a generalization toward learning additional physics, and an intriguing demonstration of a possible direction for turbulence simulation, a formulation was also considered that does not exactly enforce a discrete divergence-free constraint in LES. Rather, the physics of the pressure variable on the coarse LES grid is directly learned from the data. An advantage is reduced computational cost since it does not require a Poisson solver. 

Extending to more complex Navier--Stokes turbulence flows is an obvious direction.  An important challenge here will be to obtain and use training sets with sufficiently rich variability to enable extrapolation to new configurations.  

The DPM design is also much broader than Navier--Stokes turbulence.  We anticipate that it might be particularly suited to still more intricate sub-grid-scale modeling problems, such as in mixing, combustion, or additional physical mechanisms coupled with turbulence.  The formulation is such that experimental data could be incorporated within the formulation even if the physical description is unclear. It is designed to leverage known and resolvable physics to the highest degree possible, not replace it.

\subsection*{Acknowledgments}

This material is based in part upon work supported by the Department of Energy, National Nuclear Security Administration, under Award Number DE-NA0002374.  This research is part of the Blue Waters sustained-petascale computing project, which is supported by the National Science Foundation (awards OCI-0725070 and ACI-1238993) and the State of Illinois. Blue Waters is a joint effort of the University of Illinois at Urbana--Champaign and its National Center for Supercomputing Applications.

\appendix
\section{Additional Discretization Details}
\label{a:discreteforms}

%\subsection{Flow solver}

\subsection{Divergence-free projection of DNS fields}

Since the filtered DNS solution does not satisfy the discrete divergence-free condition $\cD^{\Delta}(\ol{\bu} ) = 0$ on the coarse LES mesh, as a pre-processing step we make the appropriate $\ell^2$ projection at each time $t$: 
\begin{equation}
\mathbf{w}_{t} = \argmin_{\mathbf{w} \in \mathcal{C}} \frac{1}{2} \norm{ \ol{ \mathbf{U} }_{t} - \mathbf{w} }_2^2, 
%&\phantom{.}& \textrm{such that} \phantom{....} \frac{u_{i, j, k} - u_{i, j-1, k}}{\Delta}  + \frac{v_{i, j, k} - v_{i-1, j, k}}{\Delta}  + \frac{w_{i, j, k} - w_{i, j, k-1}}{\Delta}  = 0,
\label{ConstrainedOptDiscrete}
\end{equation}
where $\ol{\mathbf{U}}_t =  \big{\{} \ol{ \mathbf{u} }(t, i \Delta, j \Delta, k \Delta)  \big{\}}_{i,j,k=1}^N$ is the filtered DNS data on the coarse $N^3$ grid at time $t$ and where
\begin{eqnarray}
\mathcal{C} = \bigg{\{} \mathbf{w} \in \mathbb{R}^{3 \times N \times N \times N}:  &\phantom{.}& \frac{w_{1, i+1, j, k} - w_{1, i, j, k}}{\Delta} +   \frac{w_{2, i, j+1, k} - w_{2, i, j, k}}{\Delta}  + \frac{w_{3, i, j, k+1} - w_{3, i, j, k}}{\Delta}  = 0 \phantom{....} \notag \\
&\phantom{.}& \textrm{and} \phantom{....}  w_{m, N, j, k} = w_{m, 1, j, k}, w_{m, i, N, k} = w_{m, i, 1, k}, w_{m, i, j, N} = w_{m, i, j, 1}   \bigg{\}}. \label{setC}
\end{eqnarray}
Gradients of $\mathbf{w}$ in \eqref{setC} have been derived using the same second-order, staggered-mesh central difference scheme as the flow solver~\cite{Harlow1965}.

That is, we estimate the $\mathbf{w}_t$ nearest to $\ol{ \mathbf{U} }_t$ such that $\mathbf{w}_t$ satisfies the discretized divergence-free condition on the coarse LES grid. Using the method of Lagrange multipliers, the minimizer $\mathbf{w}$ of the constrained optimization problem (\ref{ConstrainedOptDiscrete}) can be shown to satisfy 

%As for pressure-projection numerical methods, the minimizer $\mathbf{w}$ of the constrained optimization problem (\ref{ConstrainedOptDiscrete}) satisfies

\begin{eqnarray}
w_{1, i,j,k} &=& \ol{U}_{1, i,j,k}  - \frac{ \lambda_{i+1,j,k} - \lambda_{i, j, k} }{\Delta}, \notag \\
w_{2, i,j,k} &=& \ol{U}_{2, i,j,k}  - \frac{ \lambda_{i,j+1,k} - \lambda_{i, j, k} }{\Delta}, \notag \\
w_{3, i,j,k} &=& \ol{U}_{3, i,j,k}  - \frac{ \lambda_{i,j,k+1} - \lambda_{i, j, k} }{\Delta}, 
\end{eqnarray}
where the pressure-like Lagrange multiplier $\lambda$ is a solution to the discrete Poisson equation
\begin{equation}
\begin{split}
\frac{ \ol{U}_{1, i,j,k} - \ol{U}_{1, i-1,j,k} }{\Delta} & + \frac{ \ol{U}_{2, i,j,k} - \ol{U}_{2, i,j-1,k} }{\Delta}  + \frac{ \ol{U}_{3, i,j,k} - \ol{U}_{3, i,j,k-1} }{\Delta}\\ & =   \frac{ \lambda_{i,j+1,k} - 2 \lambda_{i, j, k} +  \lambda_{i, j-1, k} }{\Delta^2} 
+ \frac{ \lambda_{i+1,j,k} - 2 \lambda_{i, j, k} + \lambda_{i-1, j, k}  }{\Delta^2}  + \frac{ \lambda_{i,j,k+1} -2 \lambda_{i, j, k} + \lambda_{i,j,k-1} }{\Delta^2},
\label{DiscretePoissonEqn}
\end{split}
\end{equation}
with periodic boundary conditions. 
This yields $\mathbf{w}_t$ used for the target $\bV$ in (\ref{ObjectiveFunction}). 

%\section*{References}
%\bibliographystyle{my-elsarticle-num}
%\bibliography{library.bib}

\end{document}